\documentclass[journal=aesccq,manuscript=article,layout=twocolumn]{achemso}

\SectionNumbersOn

\usepackage{graphicx}
\PassOptionsToPackage{hyphens}{url}
\usepackage[bookmarks]{hyperref}
\usepackage[version=3]{mhchem} 
\setcitestyle{numbers,open={(},close={)}}

\author{Deniz Soysal}
\affiliation[KU Leuven]{KU Leuven, Kasteelpark Arenberg 10, 3001 Leuven, Belgium}

\author{Xabier García--Andrade}
\affiliation[KU Leuven]{KU Leuven, Kasteelpark Arenberg 10, 3001 Leuven, Belgium}

\author{Laura E. Rodriguez}
\affiliation[LPI]{Lunar \& Planetary Institute, Universities Space Research Association, 3600 Bay Area Boulevard, Houston, TX 77058, USA}

\author{Pablo Sobron}
\affiliation[Impossible Sensing]{Impossible Sensing, 20 South Sarah Street, St. Louis, MO 63108, USA}

\author{Laura M. Barge}
\affiliation[JPL]{Jet Propulsion Laboratory, California Institute of Technology, 4800 Oak Grove Drive, La Cañada Flintridge, CA 91011, USA}

\author{Renaud Detry}
\affiliation[KU Leuven, PSI]{KU Leuven, Dept.\ Electrical Engineering, Research Unit Processing Speech and Images (PSI), Kasteelpark Arenberg 10, 3001 Leuven, Belgium}
\alsoaffiliation[KU Leuven, RAM]{KU Leuven, Dept.\ Mechanical Engineering, Research Unit Robotics, Automation and Mechatronics (RAM), Celestijnenlaan 300, 3001 Leuven, Belgium}
\email{renaud.detry@kuleuven.be}

\title[Reevaluating Convolutional Neural Networks for Spectral Analysis: A Focus on Raman Spectroscopy]
  {Reevaluating Convolutional Neural Networks for Spectral Analysis: A Focus on Raman Spectroscopy}

\keywords{Deep learning, science autonomy, minerals, semi‑supervised learning, ocean worlds, astrobiology, deep-sea exploration}

\begin{document}
\sloppy

\begin{abstract}

Autonomous Raman instruments deployed on Mars rovers, deep-sea landers, and mobile field robots must interpret \emph{raw} spectra that are distorted by fluorescence baselines, peak shifts, and limited ground‑truth labels. Using rigorously documented subsets of the RRUFF mineral database, we systematically evaluate one-dimensional convolutional neural networks (CNNs) and report four practical advances:  

(i) Reproducible, baseline‑independent classification: Compact end‑to‑end CNNs surpass $k$‑nearest‑neighbors and support‑vector classifiers built on handcrafted peak features, eliminating background‑correction and peak‑picking stages. Unlike earlier work, we isolate the contribution of learned features and ensure full reproducibility by releasing all data splits and preprocessing scripts.

(ii) Pooling controlled robustness: By adjusting a single pooling parameter, CNNs accommodate Raman shift displacements up to $30 \,\mathrm{cm^{-1}}$, enabling a practical trade-off between translational invariance and class resolution, aligned with instrument stability and spectral variability.

(iii) Label-efficient learning: Semi-supervised generative adversarial networks and contrastive pretraining improve classification accuracy by up to 11\,\% when only 10\,\% of labels are available. While smaller than gains in vision tasks—due to Raman spectra’s lower complexity—these methods remain valuable for autonomous deployments with limited annotation.

(iv) Constant time adaptation: Freezing the CNNs backbone and retraining only the softmax layer transfers the model to unseen minerals at $\mathcal{O}(1)$ inference cost, outperforming Siamese networks on resource-limited robotic processors.  

The resulting workflow—\emph{train directly on raw spectra, tune pooling to instrument tolerance, add semi-supervision when labels are scarce, and fine-tune lightly for new targets}—offers a practical path toward robust, low-footprint Raman classification in autonomous field settings.

\end{abstract}

\section{Introduction}

Raman spectroscopy provides fast, non-destructive, \emph{in situ} chemical fingerprints that enable identification of minerals, organics, and biomaterials. A Raman spectrum is produced by measuring the inelastic scattering of light, where the scattered photons correspond to molecular vibrations unique to the material’s chemical structure, producing diagnostic peaks but often with weak signals and fluorescence backgrounds that complicate analysis. In the last decade, machine learning (ML) paired with spectroscopy—including Raman—has accelerated discoveries across analytical chemistry, biology, and planetary science by mining large, nonlinear spectral datasets \cite{long2022a,pomyen2020a,signoroni2019a,chen2020a,mamede2021a,sadaiappan2023a,debus2021a}. This progress reflects better computation, accessible ML libraries, and instruments that generate high-dimensional outputs (Raman, NMR, LC–MS).

\paragraph{Machine learning methods}
Spectral ML generally falls into supervised and unsupervised paradigms. Supervised learning relies on labeled datasets to train models for classification or regression tasks, proving essential for applications requiring precise predictions such as early disease diagnosis \cite{ryzhikova2021a}, forensic residue analysis \cite{banas2023a}, and microplastics monitoring \cite{lei2022a}. On the other hand, unsupervised learning exploits unlabeled data to reveal latent structure.

\paragraph{ML for autonomous exploration}
In Earth and planetary missions, supervised ML underpins scientific autonomy on robots operating from deep-sea vents to extraterrestrial terrains, where spectrometers guide real-time decisions on sampling and goal replanning \cite{barker2020a,lawrence2023a,liu2022a,hydrothermal_vents,takahashi2020a,mcmutrtry2005a,zhang2019a}. Limited storage and bandwidth force onboard prioritization of what to transmit or retain \cite{theiling2022a,poian2022a}. Emphasis on autonomy spans deep-sea mining \cite{sartore2019a}, Mars missions \cite{poian2022a,wettergreen2014a}, and planned explorations of icy ocean worlds \cite{theiling2022a,hand2018a}.

\paragraph{Challenges in applying machine learning to spectral data}
Traditional ML techniques, such as $k$-nearest neighbors (KNN) and support vector machines (SVM), typically rely on heavy preprocessing to manage high dimensionality, fluorescence baselines, and noise \cite{long2022a,debus2021a,C7AN01371J,LIU2019175}. Baseline correction is critical for Raman spectroscopy \cite{C7AN01371J}, with analogous handling for the Bremsstrahlung effect in LIBS \cite{wiens2013a} and chromatographic drift \cite{kensert2021a}. Yet expert-driven preprocessing can inject bias, and heterogeneous pipelines hinder reproducibility. A second constraint is label scarcity: obtaining mineral ground truth (XRD/chemistry) is costly and labor-intensive. Ground-truthing measurements from surface-based techniques such as Raman, LIBS, and Vis-IR spectroscopy further complicate the process, as they often require detailed comparisons between bulk and surface analyses. Biomedical and field datasets are further limited by permits, logistics, and budgets. These realities motivate label-efficient models and reproducible workflows that leverage unlabeled data. Additionally, approaches tailored to the unique characteristics of spectral data could help overcome these limitations, enabling more efficient and accurate use of both labeled and unlabeled datasets.

\paragraph{Deep learning methods}
Deep learning, a branch of ML, utilizes a series of nonlinear transformations, often implemented as neural networks, to extract meaningful insights from data. Deep learning models learn nonlinear representations directly from data and can operate on \emph{raw} spectra, reducing reliance on handcrafted preprocessing \cite{C7AN01371J}. Neural networks encompass diverse architectures but share a common structure comprising three key components: an input layer for data ingestion, one or more hidden layers where nonlinear transformations are applied, and an output layer that generates predictions. Fully connected deep neural networks (DNNs) interconnect every node between layers, but they scale poorly as input dimensionality grows. Convolutional neural networks (CNNs), by contrast, use local kernels and pooling to extract compact, hierarchical features and generalize better on images and spectral signals \cite{lecun2015deep,long2022a,chen2020a,debus2021a,C7AN01371J,berlanga2022a,xu2022a,hou2020a,shariat2022a}. CNNs' built‑in inductive biases (such as locality and translational invariance) align particularly well with the characteristics of image data and facilitate efficient feature extraction. 

\paragraph{Challenges for deep learning in spectral data}
Unlike images, small Raman peak shifts can be chemically meaningful; indiscriminate translational invariance is harmful \cite{Zhang2019TransferlearningbasedRS}. CNNs' invariance must therefore be controlled via architectural adjustments and hyperparameter tuning to be robust to instrument drift yet sensitive to class-defining displacements. Although CNNs need fewer labels than fully connected DNNs, they remain more data-hungry than KNN or SVM because they \emph{learn} features rather than rely on engineered ones. Label-efficient strategies are thus important: domain-aware augmentation \cite{C7AN01371J,blazhko2021a,wang2021a}, semi-/self-supervised learning to exploit unlabeled spectra \cite{vanEngelen2020,wu2018a}, and transfer learning to adapt pretrained backbones to new datasets \cite{Zhang2019TransferlearningbasedRS,han2021a}. Although these strategies show promise, their effectiveness in the context of Raman spectroscopy and other spectral data remains an area requiring further investigation.

\paragraph{Applications of ML in Raman spectroscopy}
Traditional Raman pipelines have used multivariate discriminant analysis \cite{chemometrics},  partial least-squares regression \cite{6120496}, KNN \cite{Li_2016}, and SVM \cite{Widjaja2008-gy}. More recently, CNNs have emerged as a more effective approach for analyzing Raman spectra \cite{10.1117/12.464039, C7AN01371J}. To address the challenge of limited labeled data, Liu \emph{et al.} \cite{LIU2019175} proposed a representation learning approach using Siamese networks \cite{Koch2015SiameseNN}, shifting from multi-class classification to binary classification. Despite progress, two gaps remain: (i) image-centric CNNs inductive biases are not fully aligned with Raman peak physics, and (ii) most methods do not fully leverage the abundant unlabeled spectra. Addressing both is essential especially in label-scarce settings.

\paragraph{Scope of This Study}

We build upon prior work by focussing on understanding the role of inductive biases in CNNs' architectures and exploring data-efficient strategies such as semi-supervised learning and transfer learning. Unlike previous studies that primarily benchmark models, our work provides an in-depth analysis of CNNs robustness to spectral shifts, investigates the complexity of low-level spectral features, and offers practical guidelines for deploying CNNs in data-scarce environments.

Using rigorously documented subsets of the RRUFF database \cite{rruff}, our objectives are fourfold:

\begin{enumerate}
    \item \textbf{Investigating CNNs Inductive Biases:} Analyze how the inherent inductive biases of CNNs' architectures impact Raman spectral classification, and propose architectural adjustments to enhance performance.
    \item \textbf{Benchmarking Against Traditional Methods:} Compare CNNs with classical machine learning approaches, including SVM and KNN, on both raw and processed Raman spectra.
    \item \textbf{Exploring semi-supervised learning:} Evaluate semi-supervised methods such as semi-supervised generative adversarial networks (SGANs) and contrastive learning to improve accuracy by leveraging unlabeled spectral data.
    \item \textbf{Analyzing low-level spectral features:} Examine the complexity and structure of low-level Raman features to understand their role in model performance and guide architecture design.
\end{enumerate}

To keep the narrative clear, we interleave methodological discussion with results. Section~\ref{sec:data} introduces our RRUFF subsets; Section~\ref{sec:traditional_methods} establishes SVM/KNN baselines; Section~\ref{sec:cnn} shows CNNs’ superiority on raw spectra; Section~\ref{sec:ind_bias} tunes inductive biases for shift robustness; Section~\ref{sec:semi-supervised} develops semi‐supervised approaches; Section~\ref{sec:low_level} investigates feature complexity; and Section~\ref{sec:transfert_learning} demonstrates lightweight transfer learning for new mineral classes.

\section{The Database} 
\label{sec:data}

Our experiments utilize the RRUFF Database\footnote{\url{https://rruff.info/}}, the largest publicly available mineral spectral database \cite{rruff}. The database currently holds spectra for 4\,216 curated specimens (covering 2\,351 mineral species out of 6\,006 IMA-approved minerals).

\paragraph{RRUFF content and provenance}
The RRUFF Project was created to provide “a complete reference set of Raman spectra, X-ray diffraction patterns and electron-microprobe analyses for every mineral species”\,\cite{rruff}. Each database record corresponds to a single, archived specimen with a unique RRUFF ID and includes (i) Raman spectra acquired at 532 nm (often additional wavelengths) in both \textit{Raw} and baseline-corrected \textit{Processed} forms; (ii) a powder-XRD pattern used to confirm phase purity; (iii) full chemical analysis and ideal formula; (iv) detailed acquisition metadata (laser power, spot size, spectrometer, crystal orientation, temperature); and (v) locality information and a photomicrograph. Because every spectrum is traceable to a specimen whose identity has been independently validated by XRD and chemistry, the mineral name supplied by RRUFF constitutes a ground-truth label for supervised learning.

\paragraph{What a spectrum represents}
A Raman entry in RRUFF is a one-dimensional vector $I(\tilde{\nu})$ of photon counts (or arbitrary intensity units) as a function of Raman shift $\tilde{\nu}$ ($\mathrm{cm^{-1}}$) measured from a single laser spot on the specimen.  Most specimens are analyzed in multiple modes:  
high-resolution scans (narrow range, $\sim70$–$1\,400 \,\mathrm{cm^{-1}}$) that resolve diagnostic peaks,  
low-resolution scans (broad range, $70$–$6\,500 \,\mathrm{cm^{-1}}$) that capture lattice modes and overtones, and  optionally oriented measurements with the crystallographic axes aligned to the laser polarization.

\paragraph{Terminology used in the remainder of this paper}
\begin{itemize}
  \item \textbf{Spectrum}: a single Raman vector $I(\tilde{\nu})$ as defined above.  
  \item \textbf{Labeled spectrum}: a spectrum whose mineral species (e.g.\ “quartz”) is provided to the model during training or evaluation; in practice this means the species tag supplied by RRUFF.  
  \item \textbf{Unlabeled spectrum}: a spectrum for which the species tag is withheld, either deliberately (e.g. to create semi-supervised splits) or because ground-truth is absent in a field-deployment scenario.  
\end{itemize}

We will consistently use the terms labeled and unlabeled in this sense throughout the paper. For every Raman measurement, RRUFF stores a text file whose file name indicate the processing state—ending in “…\_Raman\_Data\_\textbf{RAW}.txt” or “…\_Raman\_Data\_\textbf{Processed}.txt”.  Throughout this paper we refer to these two sets simply as

\begin{itemize}
  \item \textbf{RRUFF-raw}: spectra as recorded in the RAW format, without baseline preprocessing;
  \item \textbf{RRUFF-clean}: baseline-corrected spectra (processed format, where baselines have been removed by \textsc{CrystalSleuth}\cite{rruff}).
\end{itemize}

\subsection{Subset of the RRUFF database used in this study}
\label{sec:subset_data}
We consider two subsets of the RRUFF database: 

\begin{itemize} 
    \item \textbf{Low-resolution unoriented subset:} Downloaded from \url{https://rruff.info/zipped_data_files/raman/LR-Raman.zip}, containing unoriented low-resolution samples. Classes with fewer than eight samples are pruned, and only the RRUFF-raw spectra are retained.
    \item \textbf{High-resolution oriented subset:} Downloaded from \url{https://rruff.info/zipped_data_files/raman/excellent_oriented.zip}, containing oriented high-resolution samples. Same pruning is applied, and only the RRUFF-clean spectra are retained.
\end{itemize}

This results in two datasets: one composed of RRUFF-raw spectra and another of RRUFF-clean spectra. These datasets are then split into training and test sets. The specific subsets used in our experiments can be downloaded from \url{https://github.com/denizsoysal/Raman_spectra_data}.

\subsection{Preprocessing}
\label{sec:preprocessing}

In addition to the baseline correction performed in the RRUFF-clean spectra by the database maintainers, we apply our own preprocessing pipeline systematically to both RRUFF-raw and RRUFF-clean data. These steps are fully automated and applied in batch mode across the entire dataset, without any human intervention at the sample level.

\paragraph{Class Pruning} 
Classes with fewer than $n_{\text{min}} = 8$ samples are removed to maintain sufficient training data per category.

\paragraph{Wavelength Range Selection} 
Raman shifts outside the $(200, 1600)\,\mathrm{cm^{-1}}$ range are discarded. This decision is motivated by the observation that peaks are concentrated in the lower Raman shift values, whereas higher shifts provide limited discriminative information. Additionally, the lowest Raman shift region corresponds to intermolecular vibrations, which are not discriminative for mineral classification. Based on this observation, the models are restricted to the $\nu \in (200 ,1600)\,\mathrm{cm^{-1}}$ domain.

\paragraph{Interpolation and Resampling} 
To standardize input dimensions and spectral resolution, each raw spectrum’s shift–intensity pairs are projected onto a common Raman‐shift grid spanning $200$-$1600\,\mathrm{cm^{-1}}$ in $1\,\mathrm{cm^{-1}}$ steps. These bounds and step size are chosen based on statistical summaries of all spectra, including minimum/maximum shift ranges and the average sampling resolution. This yields $1401$ points, which we then truncate to $1392$ points (the nearest lower multiple of 16) to satisfy GPU alignment constraints. Projection is performed via linear interpolation, with any values outside a spectrum’s original range set to zero. No additional smoothing or filtering is applied, preserving spectral fidelity. The resulting $1392$‑dimensional intensity vector ensures uniform resolution, consistent dimensionality, and efficient batching for neural network training.

\paragraph{Normalization} 
All spectra are normalized to ensure consistent intensity scaling, reducing variability introduced by acquisition conditions.

\vspace{1\baselineskip}

We emphasize that the preprocessing steps described above are computationally inexpensive and applied in batch mode across the entire dataset. No human intervention is required on a per-sample basis, which ensures reproducibility and mitigates the risks of manual bias typically associated with expert-guided preprocessing.

\subsection{Final Dataset}

After pruning classes with fewer than eight spectra and applying the
pre-processing pipeline, the corpus splits into two subsets:

\begin{itemize}
  \item \textbf{RRUFF-clean}: 3 425 spectra, 155 mineral classes  
        (mean 22.1 spectra per class).
  \item \textbf{RRUFF-raw}: 1 311 spectra, 102 mineral classes  
        (mean 12.9 spectra per class).
\end{itemize}

\begin{figure*}[t]
  \centering
  \begin{minipage}[t]{0.475\textwidth}
    \centering
    \includegraphics[width=\textwidth]{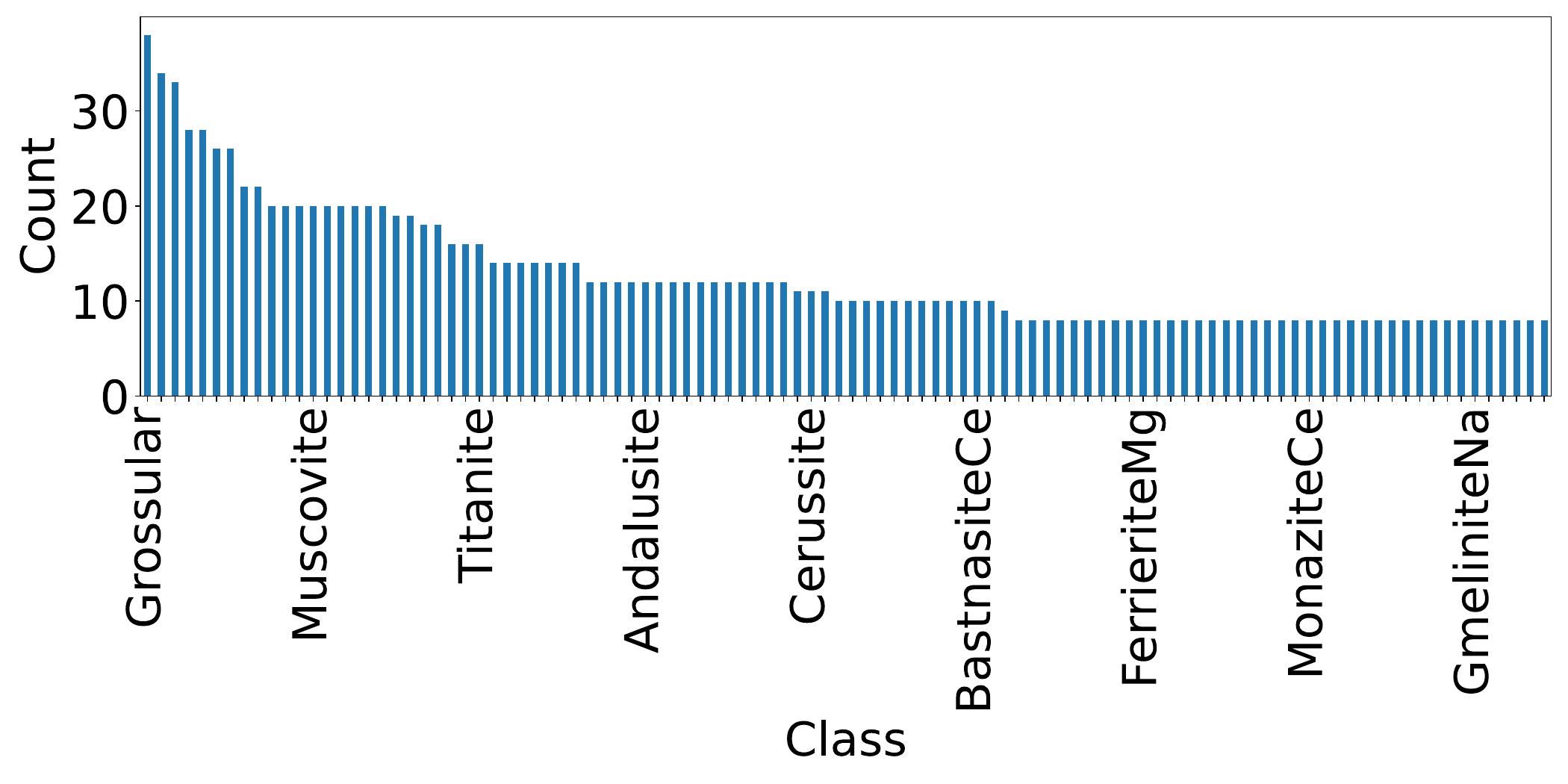}
    \caption*{(a) Class-size histogram for the RRUFF-raw subset.}
  \end{minipage}%
  \hfill
  \begin{minipage}[t]{0.475\textwidth}
    \centering
    \includegraphics[width=\textwidth]{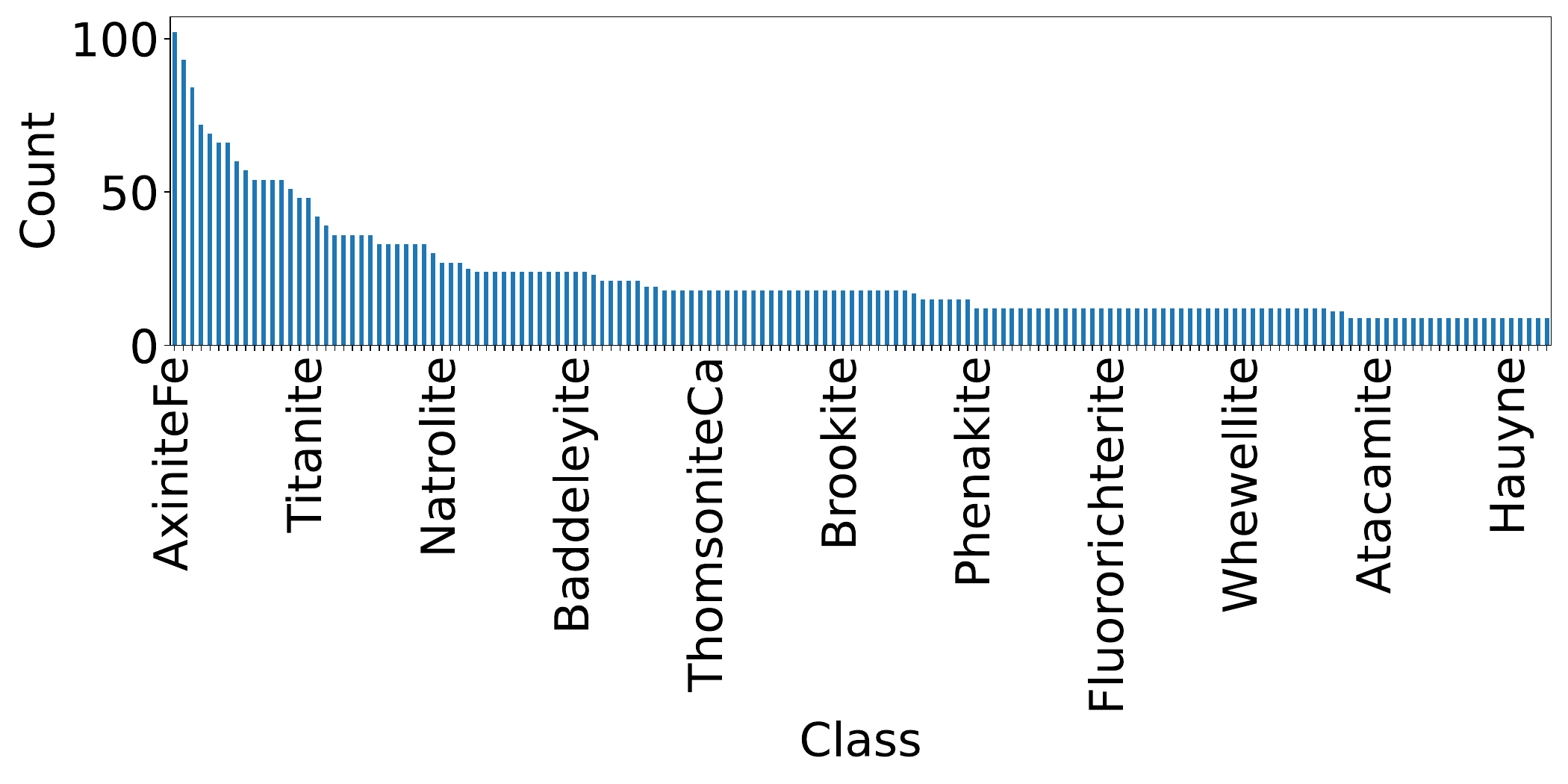}
    \caption*{(b) Class-size histogram for the RRUFF-clean subset.}
  \end{minipage}
  \caption{Number of spectra per mineral class after all pre-processing steps.
  The imbalanced distribution motivates the use of class-weighted loss in model
  training.}
  \label{fig:samples_per_class}
\end{figure*}

Both subsets are highly imbalanced (Figure~\ref{fig:samples_per_class}), with
most species represented by fewer than 20 spectra.  
To mitigate this, we apply class weights in the cross-entropy loss, forcing the network to pay proportionally more attention to under-represented classes.

\subsection{Illustrative spectral variability}

\begin{figure*}[t]
    \centering
    \begin{minipage}[t]{0.475\textwidth}
        \centering
        \includegraphics[width=1\textwidth]{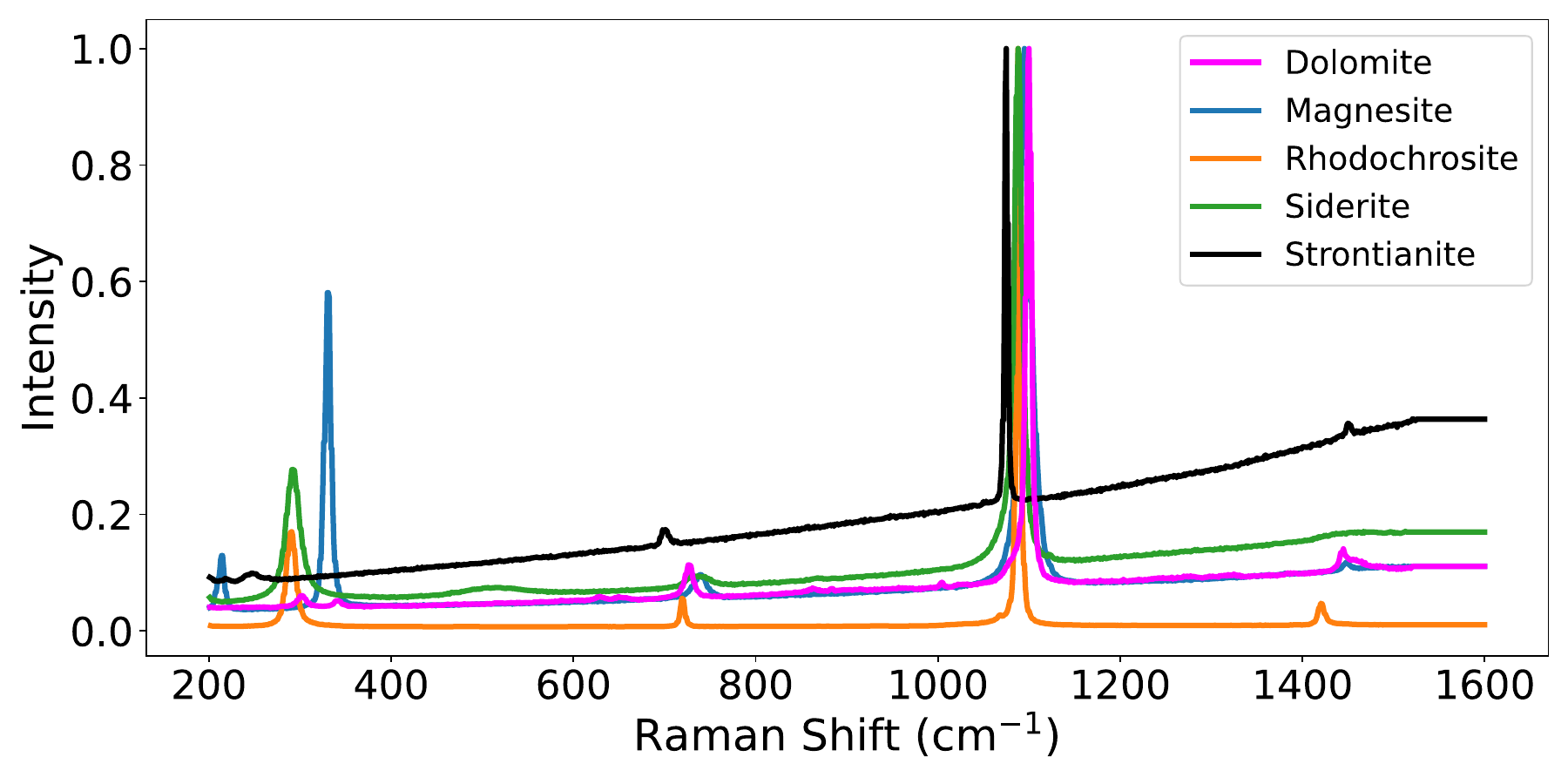}
        \caption*{(a) RRUFF-raw spectra, showing fluorescence baseline.}
    \end{minipage}
    \hfill
    \begin{minipage}[t]{0.475\textwidth}
        \centering
        \includegraphics[width=1\textwidth  ]{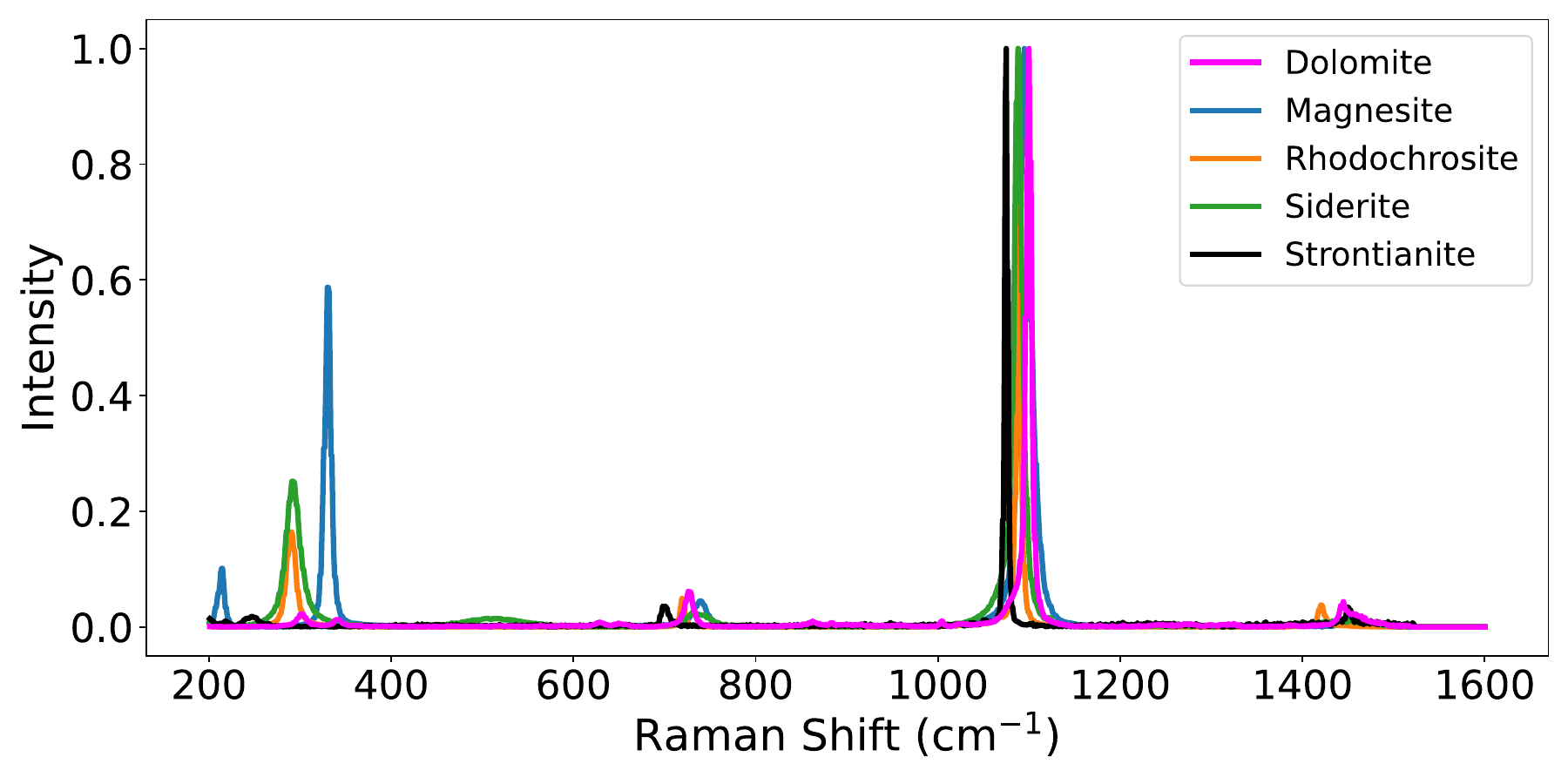}
        \caption*{(b) RRUFF-clean spectra, with baseline removed.}
    \end{minipage}
    \caption{RRUFF-raw and  RRUFF-clean spectra for dolomite (\ce{CaMg(CO3)2}) plotted alongside four compositionally related carbonate minerals—magnesite (\ce{MgCO3}), rhodochrosite (\ce{MnCO3}), siderite (\ce{FeCO3}), and strontianite (\ce{SrCO3}).  The shared carbonate $\nu_1$ symmetric stretch at $1 080$–$1 100\,\mathrm{cm^{-1}}$ and lattice modes near $300\,\mathrm{cm^{-1}}$ illustrate the high intra-group spectral similarity, while the RRUFF-raw scans highlight fluorescence backgrounds removed in the RRUFF-clean scans.}
    \label{fig:rruff_sample_dolomite_and_friend}
\end{figure*}

\begin{figure*}[t]
    \centering
    \begin{minipage}[t]{0.475\textwidth}
        \centering
        \includegraphics[width=1\textwidth]{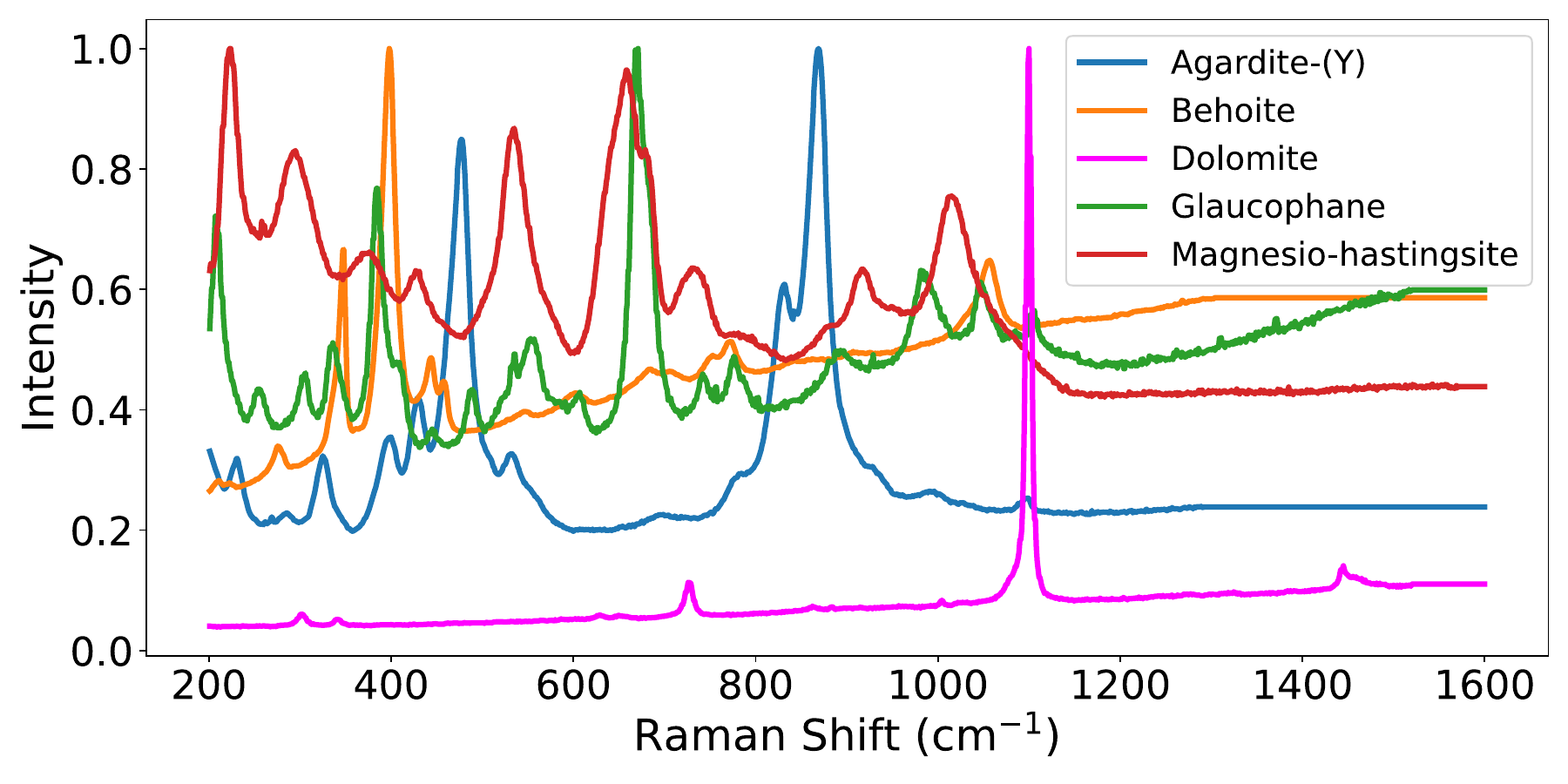}
        \caption*{(a) RRUFF-raw spectra, showing fluorescence baseline.}
    \end{minipage}
    \hfill
    \begin{minipage}[t]{0.475\textwidth}
        \centering
        \includegraphics[width=1\textwidth  ]{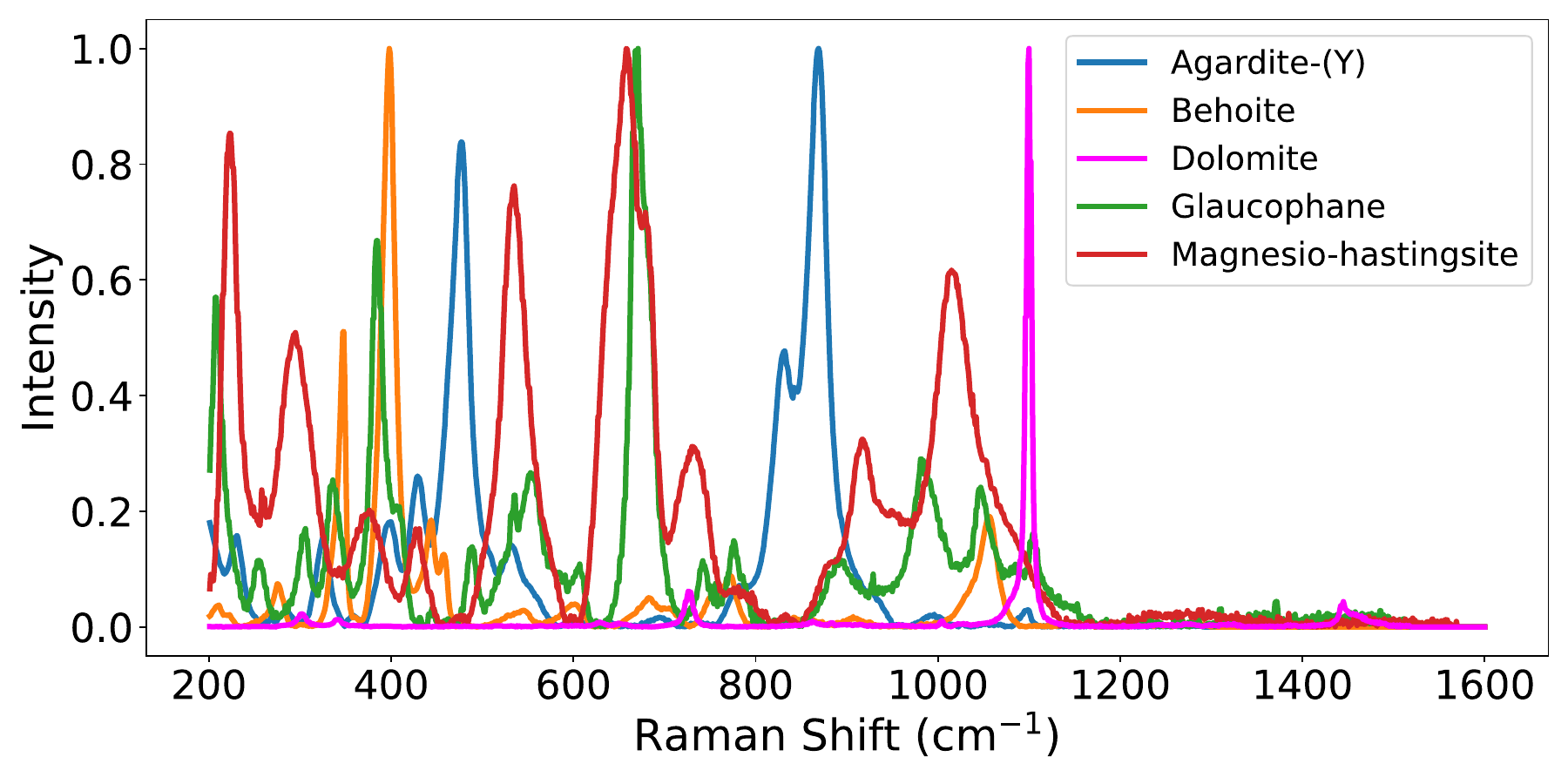}
        \caption*{(b) RRUFF-clean spectra, with baseline removed.}
    \end{minipage}
    \caption{RRUFF-raw and RRUFF-clean spectra for dolomite (\ce{CaMg(CO3)2}) plotted alongside four spectrally dissimilar minerals: agardite-(Y) (\ce{YCu6(AsO4)3(OH)6*3H2O}), behoite (\ce{Be(OH)2}), glaucophane (\ce{Na2(Mg3Al2)Si8O22(OH)2}), and magnesio-hastingsite (\ce{NaCa2(Mg4Fe)(Si6Al2)O22(OH)2}).  The dolomite trace shows the characteristic carbonate $\nu_1$ band at $1 080$–$1 100\,\mathrm{cm^{-1}}$, whereas the comparison minerals exhibit dominant arsenate or silicate stretches and entirely different lattice modes.  Together with Figure~\ref{fig:rruff_sample_dolomite_and_friend}, this panel demonstrates the spectral variability in the dataset, from very closely related classes that differ by subtle peak shifts to minerals that are spectrally distant.}
    \label{fig:rruff_sample_dolomite_and_ennemies}
\end{figure*}

Figure \ref{fig:rruff_sample_dolomite_and_friend} plots dolomite with four other carbonate minerals whose Raman signatures are almost super-imposable. All five spectra feature the characteristic carbonate band at $\sim1085\,\mathrm{cm^{-1}}$; the principal peaks differ by less than $10\,\mathrm{cm^{-1}}$, creating a fine‑resolution challenge for any classifier. Figure \ref{fig:rruff_sample_dolomite_and_ennemies} instead compares dolomite with four minerals from chemically different groups (arsenates, hydroxides, silicates). Here, the spectra share no dominant bands, illustrating markedly different vibrational chemistries.
Taken together, the two figures span the extremes of our database, from near-duplicate spectra to entirely distinct signatures.

\subsection{Data augmentation strategy}
\label{sec:data_augmentation}

Given the characteristics of Raman spectra, augmentation techniques must be carefully designed to preserve spectral peaks. We employ transformations that retain the integrity of Raman features, as detailed in Section~\ref{sec:contrastive_learning}. Data augmentation is applied in Section~\ref{sec:training_cnn} to evaluate the performance of CNNs on augmented data and in Section~\ref{sec:contrastive_learning}  in the context of semi-supervised learning.

\subsection{Implementation and Training Environment}

All deep-learning models were implemented in PyTorch. Classical machine-learning baselines (SVM and KNN) were trained with scikit-learn. Training was performed on a laptop equipped with an NVIDIA RTX~3070 GPU. Typical training time per data split was \(\sim\)1–2 hours for deep-learning experiments.

\section{Classification using SVM and KNN}
\label{sec:traditional_methods}

Before introducing advanced neural network-based approaches, it is essential to establish a performance baseline using traditional machine learning methods for Raman spectral classification. Support vector machines (SVM) and k-nearest neighbors (KNN) are two widely used techniques. SVM operates by finding an optimal hyperplane that separates data points into distinct classes while maximizing the margin between them. KNN, in contrast, is a distance-based approach that assigns an unknown data point \(x\) to the most common class among its \(k\) nearest neighbors in the dataset \(X\). (Refer to Supporting Information for further details on SVM and KNN.)

In this section, we implement both SVM and KNN classifiers not only to explore their performance but, more importantly, to serve as explicit baselines for our subsequent experiments with deep learning models (see Section~\ref{sec:cnn}). Establishing these baselines allows us to rigorously quantify the improvements introduced by convolutional neural networks and to contextualize our findings within existing machine learning frameworks for spectral data.

Prior work by Widjaja \emph{et al.}~\cite{Widjaja2008-gy} demonstrated the application of SVM for Raman spectral classification, employing preprocessing techniques such as dimensionality reduction to mitigate noise and baseline effects from fluorescence. In this study, we implement both SVM and KNN classifiers and evaluate their performance using feature extraction techniques tailored for Raman spectra. Our findings indicate that while SVM and KNN achieve satisfactory results on RRUFF-clean data, their performance degrades significantly when applied to RRUFF-raw spectral data.

\subsection{Methodology}

Feature extraction plays a crucial role in determining the effectiveness of machine learning classifiers. To ensure high classification performance, the extracted features must be highly informative regarding class distributions.

As detailed in the Supporting Information, the primary features relevant to mineral classification in Raman spectroscopy are the \textit{positions of spectral peaks}. To extract these peaks, we evaluated two peak detection methods:

\begin{enumerate}
    \item \textbf{Wavelet-based peak detection:} Uses a continuous wavelet transform to identify spectral peaks by matching the signal shape with a predefined wavelet function (\texttt{signal.find\_peaks\_cwt} from SciPy \cite{2020SciPy-NMeth}).
    \item \textbf{Local maxima detection:} Identifies peaks by comparing each value with its neighboring points (\texttt{signal.find\_peaks} from SciPy).
\end{enumerate}

We performed a grid search to determine the optimal parameters for each method. Table~\ref{tab:peak_detection} presents the classification accuracy of KNN using each peak detection approach. The wavelet-based method consistently outperformed the local maxima approach, leading us to select it as our primary peak detection technique.

\begin{table*}[t]
\captionsetup{justification=centering}
\begin{center}
\def\arraystretch{1.5}
\resizebox{\textwidth}{!}{
\begin{tabular}{|c|c|c|}
\hline
Method & Wavelet-Based & Local Maxima-Based \\
\hline
Function & \texttt{find\_peaks\_cwt} & \texttt{find\_peaks} \\
\hline
Peak detection strategy & Shape-matching using Ricker wavelets & Intensity-based local maxima \\
\hline
Best Parameters & Width: [10, 20] \newline Wavelet: Ricker & Width: 10 ~~ Prominence: max(intensity)/len(signal) \\
\hline
Top-1 Accuracy (KNN) & 70\% & 59\% \\
\hline
\end{tabular}
}
\end{center}
\caption{Comparison of peak detection methods for KNN classification. Wavelet-based detection, using shape-matching with Ricker wavelets, achieves higher accuracy than local maxima, demonstrating its robustness to spectral noise and baseline variations.}
\label{tab:peak_detection}
\end{table*}

\begin{figure*}[t]
\centering
\includegraphics[width=0.6\textwidth, trim=0 0 0 20, clip]{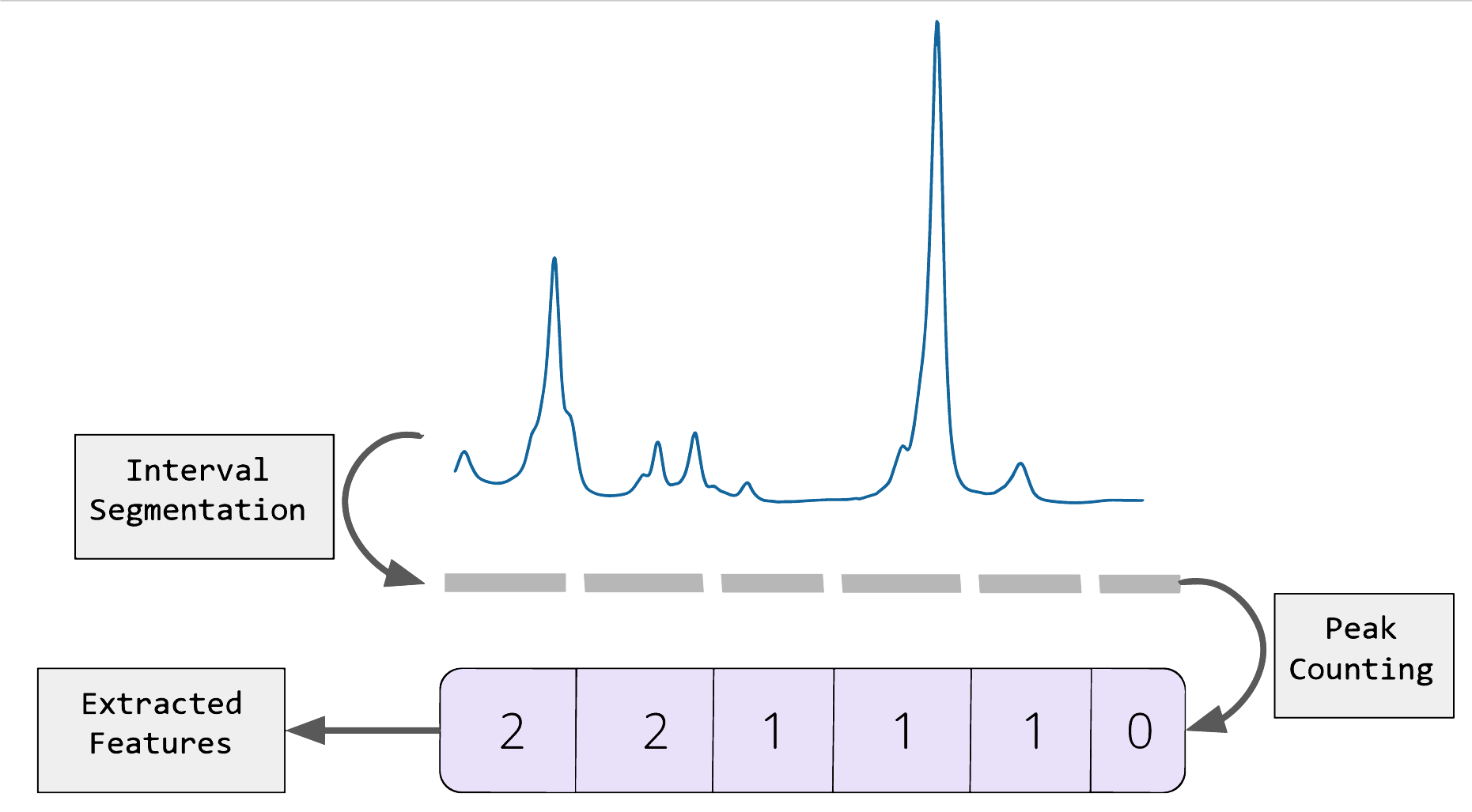}
\caption{Feature extraction process for traditional ML methods. The Raman spectrum is segmented into intervals and peaks are counted within each interval.}
\label{fig:featureex}
\end{figure*}

\paragraph{Feature vector construction}
The wavelet detector returns a set of Raman–shift positions at which peaks
occur. Because each spectrum contains a different number of peaks, we convert that variable-length list into a fixed-length feature vector by binning the shift axis and counting how many peaks fall inside each bin. We adopt a bin width of $12\,\mathrm{cm^{-1}}$, guided by the typical variability of the carbonate $\nu_1$ symmetric‐stretch band (centered near $1085\,\mathrm{cm^{-1}}$), which shifts by up to $\sim10\,\mathrm{cm^{-1}}$ between closely related species. This choice is therefore narrow enough to distinguish those fine spectral differences yet broad enough to keep the feature space compact and avoid the curse of dimensionality associated with very high-dimensional feature spaces. Given the input vector of 1392 dimensions, partitioning it into $12\,\mathrm{cm^{-1}}$ intervals produces exactly 116 bins; each spectrum is
therefore encoded as a 116-dimensional peak-count vector, independent of the number of detected peaks. Figure~\ref{fig:featureex} illustrates the procedure; the resulting vectors serve as inputs to the KNN and SVM classifiers.

\subsection{Experiments}
\label{sec:result_knn_svm}

To determine optimal hyperparameters, we performed a grid search for both KNN and SVM. The best-performing configurations were:

\begin{itemize}
    \item \textbf{KNN}: \( k=1 \), Euclidean distance\footnote{Using Euclidean distance in the
    116‑dimensional peak‑count space, the mean intra‑/inter‑class distances are
    2.2 / 6.1 for RRUFF‑clean and 6.5 / 13.1 for RRUFF‑raw. As intra‑class variation is low relative to inter‑class variation, consulting more than one neighbor adds little useful information and can even mix in other classes; hence \(k = 1\) is optimal.}
    \item \textbf{SVM}: Regularization parameter \( C=10 \), kernel: Radial Basis Function (RBF), kernel coefficient \( \gamma = 0.01 \).
\end{itemize}

Considering the class imbalance in the dataset, we employed stratified five-fold cross-validation and reported the Top-1 accuracy (fraction of spectra whose highest‑probability predicted class matches the ground‑truth label). Table~\ref{tab:Traditional_methods_results} summarizes the results.

\begin{table}[t]
\captionsetup{justification=centering}
\begin{center}
\def\arraystretch{1.5}
\resizebox{0.475\textwidth}{!}{
\begin{tabular}{|c|c|c|}
\hline
Dataset              & RRUFF-raw   & RRUFF-clean  \\ \hline
Top-1 Accuracy of SVM (\%) & 49 & 72   \\ \hline
Top-1 Accuracy of KNN (\%) & 51 & 70  \\ \hline
\end{tabular}
}
\end{center}
\caption{Performance of KNN and SVM on RRUFF-raw and RRUFF-clean Raman spectra. Both models show accuracy improvements on RRUFF-clean data.}
\label{tab:Traditional_methods_results}
\end{table}

\begin{figure*}[htbp]
\centering
\includegraphics[width=0.5\textwidth,trim={0 3cm 8cm 0}]{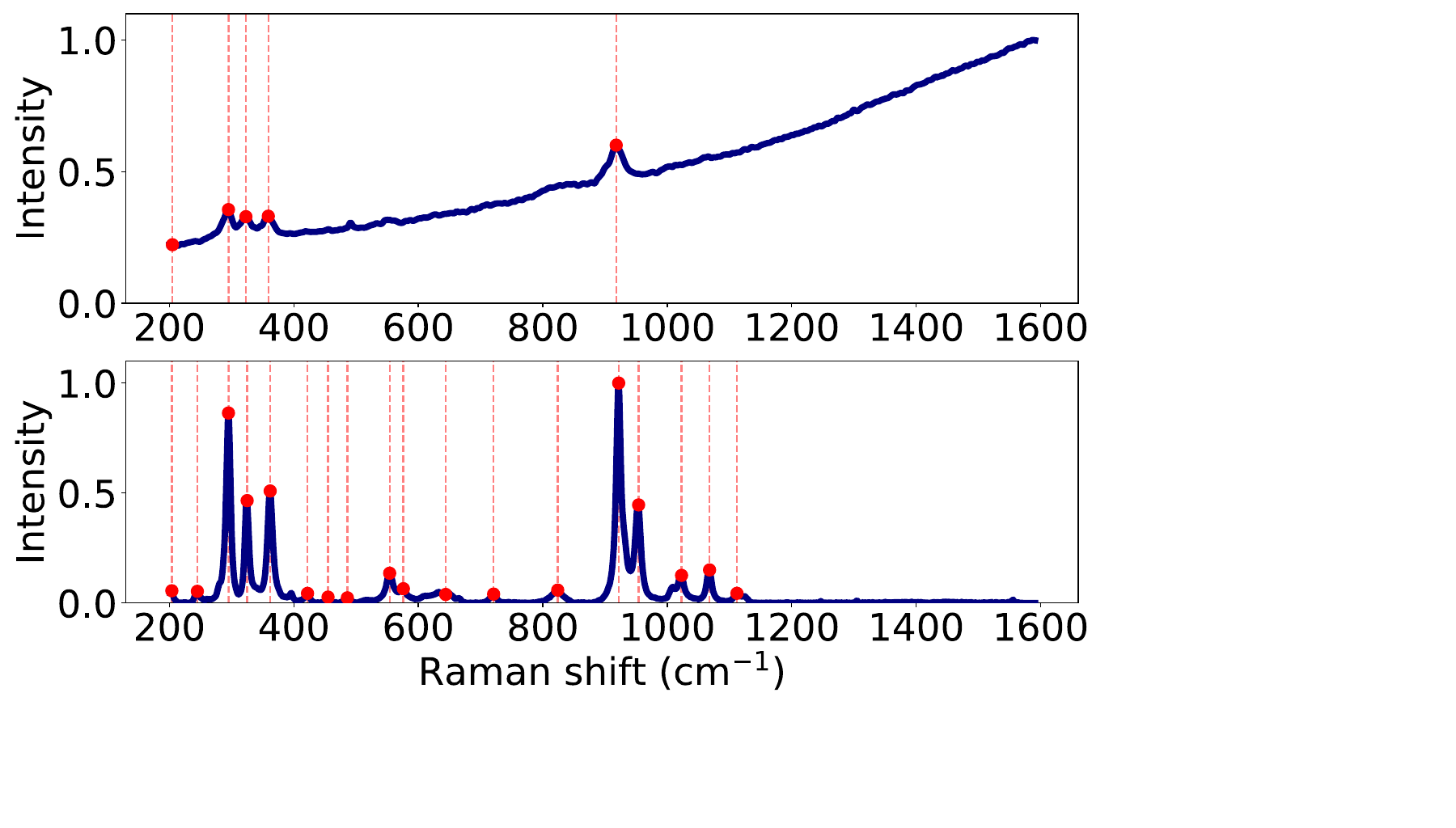}
\caption{Wavelet peak detection on Andalusite. Upper: RRUFF-raw spectrum—several diagnostic peaks are missed due to fluorescence baseline. Lower: RRUFF-clean spectrum—more peaks are detected, illustrating why peak‑based feature extraction works better on the cleaned data.}
\label{fig:peak_detection_andalusite}
\end{figure*}

Both models exhibit significantly lower accuracy on RRUFF-raw spectra compared to RRUFF-clean spectra. Our findings highlight the critical role of feature extraction in traditional ML-based Raman spectral classification. When applied to RRUFF-clean spectra, where preprocessing steps have mitigated baseline effects and noise, both KNN and SVM achieve relatively high accuracy (around 70\%). However, for RRUFF-raw spectra, these classifiers struggle due to the reliance on explicit peak detection, which becomes unreliable under noise and baseline distortions. Figure~\ref{fig:peak_detection_andalusite} illustrates this: the fluorescence baseline hides several diagnostic peaks in the RRUFF-raw spectrum (upper panel) that are recovered in the RRUFF-clean spectrum (lower panel).

An open question remains as to whether the observed performance drop is primarily due to the limitations of feature extraction or the classifiers' decision strategies within the feature space. We address this question in Section~\ref{sec:training_cnn}, where we explore whether deep learning models can overcome these challenges by learning more robust features directly from the raw spectral data.

\section{Classification using CNNs}
\label{sec:cnn}

As discussed in Section \ref{sec:traditional_methods}, traditional machine learning methods struggle to classify raw Raman spectra effectively. To overcome these limitations, we explore deep learning-based approaches. Specifically, we investigate whether convolutional neural networks (CNNs) can overcome the bottlenecks identified with classical classifiers—namely, the reliance on explicit feature extraction and potential limitations in decision strategies within the feature space. CNNs are a class of deep learning models primarily used for image processing but have also been successfully applied to spectral data. CNNs extract features by convolving a learnable filter (or kernel) with the input signal. Unlike classical convolutional operations, where the filter is predefined, CNNs learn these filters autonomously during training (refer to Supporting Information for more details on CNNs).

Although CNNs are typically designed for 2-D image data, in this work, we employ 1-D CNNs, which is well suited for 1-D data, including spectral data. 
Unlike 2-D CNNs that operate across two spatial dimensions, 1-D CNNs perform convolutions along a single axis.

\begin{figure*}[t]
\centering
\includegraphics[width=1.0\textwidth]{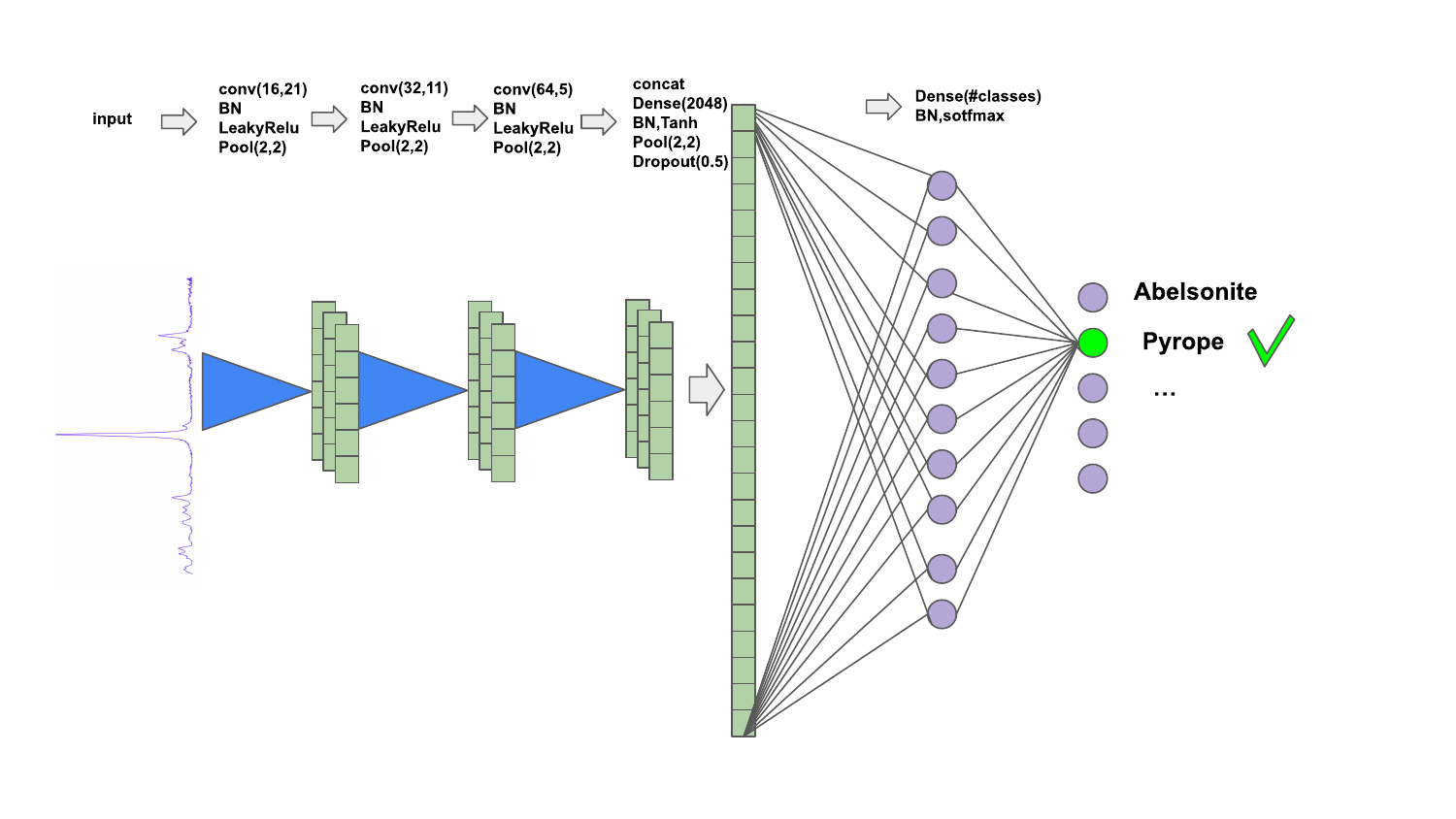}
\caption{Architecture of the 1-D CNNs model used for Raman spectral classification. 
The input consists of raw Raman spectra, which pass through three convolutional layers (16, 32, and 64 filters, with kernel sizes of 21, 11, and 5, respectively). Each convolutional layer is followed by Batch Normalization (BN), LeakyReLU activation, and max pooling (2×2). 
The extracted features are concatenated and processed through a dense layer (2048 neurons) with hyperbolic tangent (Tanh) activation, dropout (0.5), and additional pooling. The final classification layer applies a softmax function to assign the spectrum to one of the mineral classes. 
This architecture is based on the design proposed by Liu \emph{et al.}~\cite{C7AN01371J}.}
\label{fig:liuarch}
\end{figure*}

CNNs, by learning local patterns and compositional hierarchies, offer a natural advantage in processing the structured but noisy signals of Raman spectra. Liu \emph{et al.}~\cite{C7AN01371J} demonstrated the effectiveness of deep CNNs for Raman spectral classification, showing that CNNs-based models can learn automatic baseline corrections that are more robust than traditional manual preprocessing. Moreover, the hierarchical nature of CNNs enables them to learn intermediate feature representations, leading to superior classification performance compared to SVMs, particularly for datasets with a large number of classes. The CNNs' architecture used by Liu \emph{et al.}, inspired by the LeNet model~\cite{726791}, is illustrated in Figure~\ref{fig:liuarch}.

Their study, conducted on the RRUFF database, reported that CNNs outperform conventional classifiers. However, the paper lacks details regarding the specific subsets of the dataset and the preprocessing techniques applied. To provide a reproducible baseline for future research, we replicate their approach while explicitly documenting the dataset partitions and preprocessing steps in Section~\ref{sec:data}. Our objective is to enhance reproducibility and enable fairer performance comparisons.

\subsection{Methodology}

We adopt the CNNs' architecture proposed by Liu \emph{et al.}, depicted in Figure~\ref{fig:liuarch}, with one key modification: the removal of Batch Normalization layers, as preliminary experiments indicated improved performance without them. To assess the effectiveness of CNNs, we compare their performance against a Multilayer Perceptron (MLP) and KNN classifiers. The MLP model shares the same final two layers as the CNNs, but its first three convolutional layers are replaced with fully connected layers; we evaluate three capacities: MLP-small (16, 32, 64 neurons), MLP-mid (32, 64, 128 neurons), and MLP-large (64, 128, 256 neurons).

We evaluate classification performance using the following approaches:
\begin{enumerate}
    \item CNNs for end-to-end classification.
    \item CNNs for feature extraction, followed by KNN for classification.
    \item MLP for classification. 
    \item KNN for classification using handcrafted features.
\end{enumerate}

To transition from approach (1) to approach (2), we remove the final classification layer from the CNNs and use the extracted features as input for a KNN classifier. This setup allows us to further analyze insights from Section~\ref{sec:result_knn_svm} and determine whether KNN's limitations result from its inability to extract meaningful features or from intrinsic weaknesses in the classification algorithm itself.

\subsection{Experiments}
\label{sec:training_cnn}

\begin{figure*}[htbp]
\centering
\includegraphics[width=1.0\textwidth]{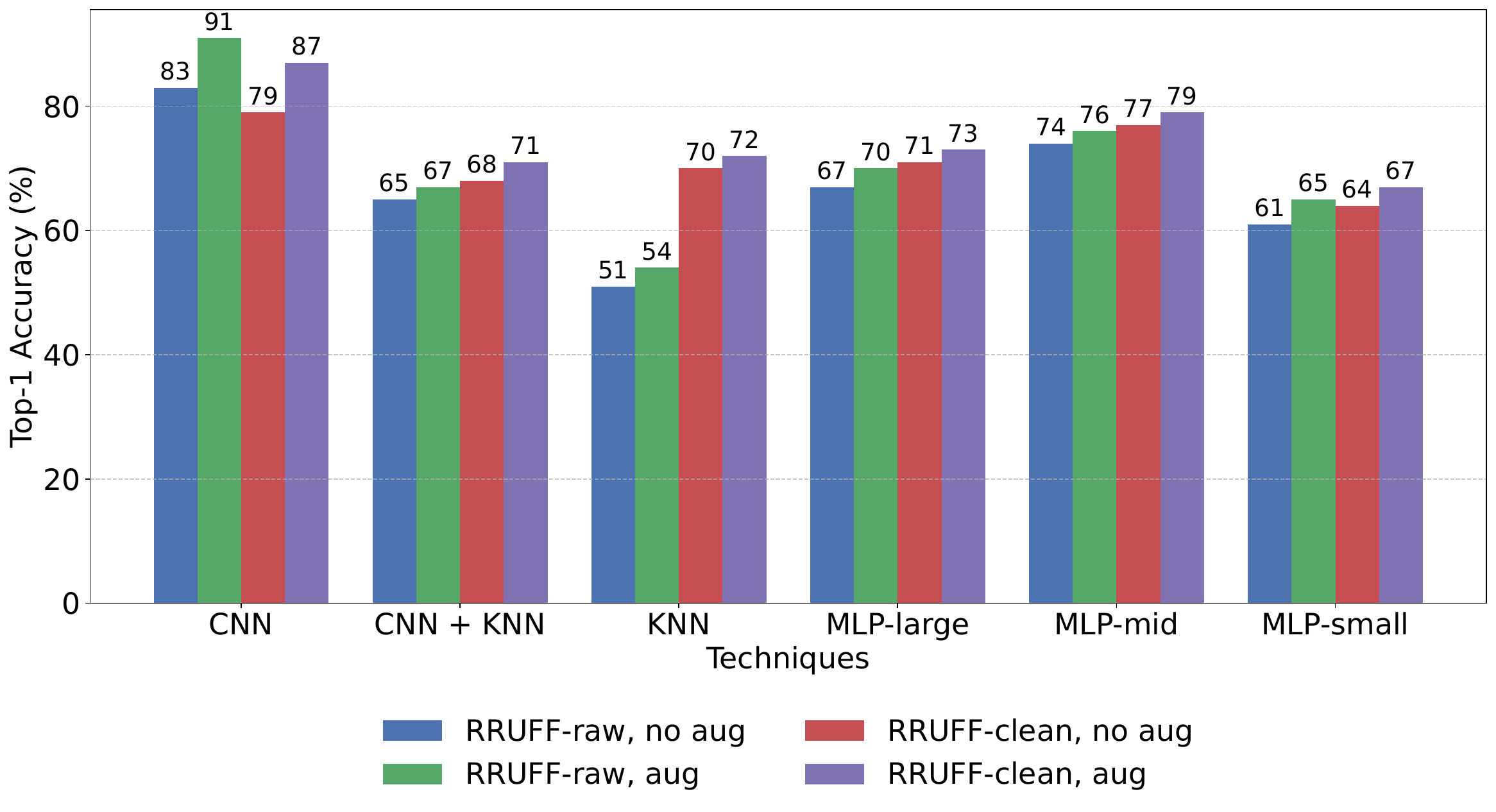}
\caption{Top-1 accuracy on the test set for CNNs, CNNs+KNN (feature extraction with CNNs followed by KNN classification), MLP, and KNN classifiers. Results are shown for both RRUFF-raw and RRUFF-clean Raman datasets, with and without data augmentation. CNNs achieve the highest performance across all settings, especially with data augmentation.}
\label{tab:CNN_accuracy}
\end{figure*}

To evaluate the performance of CNNs for Raman spectral classification, we conducted experiments using a stratified five-fold cross-validation strategy, ensuring balanced class distributions across training and validation sets, as done in Section~\ref{sec:result_knn_svm}.

The CNNs model was trained with an initial learning rate of 0.001, which was dynamically adjusted based on validation performance. If the validation loss did not improve for three consecutive epochs, the learning rate was reduced by a factor of 0.7. Additionally, early stopping was applied to prevent overfitting, terminating training if no improvement was observed for five consecutive epochs.

Figure~\ref{tab:CNN_accuracy} presents the Top-1 classification accuracy for each method. We observed that:
\begin{itemize}
    \item CNNs outperform all other approaches, achieving the highest accuracy on both RRUFF-raw and RRUFF-clean datasets.
    \item CNNs perform better on RRUFF-raw data than on RRUFF-clean data, likely due to their ability to learn robust representations and correct baseline variations automatically~\cite{C7AN01371J}.
    \item Data augmentation improves performance, as indicated by increased accuracy across all methods when augmentation was applied (see Section~\ref{sec:data_augmentation}).
    \item CNNs-extracted features improve KNN classification, but CNNs+KNN still underperforms compared to the fully end-to-end CNNs.
\end{itemize}

This suggests that: 

\begin{itemize}
    \item[(a)] CNNs demonstrate superior feature extraction capabilities compared to traditional approaches such as wavelet-based peak detection and peak counting. This is evident from the improved classification performance when using CNNs as a feature extractor, indicating that CNNs can learn more complex and relevant spectral representations. The ability of CNNs to learn hierarchical features directly from raw spectra also explains the poor performance of KNN and SVM, which rely on manually engineered features.
    
    \item[(b)] The lower classification performance of KNN compared to CNNs and MLP suggests that KNN struggles with high-dimensional feature spaces where distance metrics become less discriminative. CNNs and MLPs, on the other hand, leverage learned feature representations to improve class separability. Furthermore, the performance gap between methods (2) and (4) indicates that using CNNs for feature extraction provides a more informative representation than handcrafted features, but KNN remains inherently limited in classification capability.
\end{itemize}

\paragraph{Learning curves(training and validation loss)}
We examined the training and validation loss curves (Supporting Information, Figures~S9–S16) to assess the learning behavior of the deep learning models. On both datasets, MLP-small shows persistently high training and validation losses with slow decrease (underfitting; Figures~S9,S10), whereas MLP-large exhibits a rapid drop in training loss and a widening train–validation gap (overfitting; Figures~S13,S14). Across MLP settings, MLP-mid provides the most stable behavior, with convergent losses and a small training/validation loss gap (Figures~S11,S12). CNNs achieve the lowest validation loss and fastest convergence on both RRUFF-raw and RRUFF-clean (Figures~S15,S16).

\paragraph{Confidence and Calibration}

For CNNs, we quantified decision certainty via the \emph{confidence gap} (top--1 minus top--2 softmax probability per spectrum). CNNs exhibit well-calibrated margins, with mean confidence gaps of $0.46$ on RRUFF-raw and $0.38$ on RRUFF-clean. These values reflect the inherent difficulty of the task: the dataset contains spectrally similar minerals, such as those in Figure~\ref{fig:rruff_sample_dolomite_and_friend}. The model’s measured uncertainty on a dataset containing such closely related species is scientifically appropriate, as excessively large confidence would suggest overfitting to training artifacts rather than genuine discriminative learning. The slightly higher gap for raw spectra (0.46 vs. 0.38) parallels their superior classification accuracy, consistent with the idea that retaining full spectral information enables more decisive yet still properly calibrated predictions.

\subsubsection{Interpreting CNNs decisions with Grad-CAM}

To probe where CNNs find evidence for each mineral, we visualized their internal activation maps with Gradient-weighted
Class Activation Mapping (Grad-CAM) \cite{8237336}. For any input spectrum, Grad-CAM back-propagates the prediction score to the last convolutional layer, weights the resulting feature maps by the class-specific gradients, and collapses them to a one-dimensional \textit{importance curve}. We plot that curve as a semi-transparent color bar: warm hues highlight
Raman shifts that increase the class score, cool tones those that contribute little.

\begin{figure}[htbp]
\centering
\includegraphics[width=0.475\textwidth,trim={0 2cm 8cm 0}]{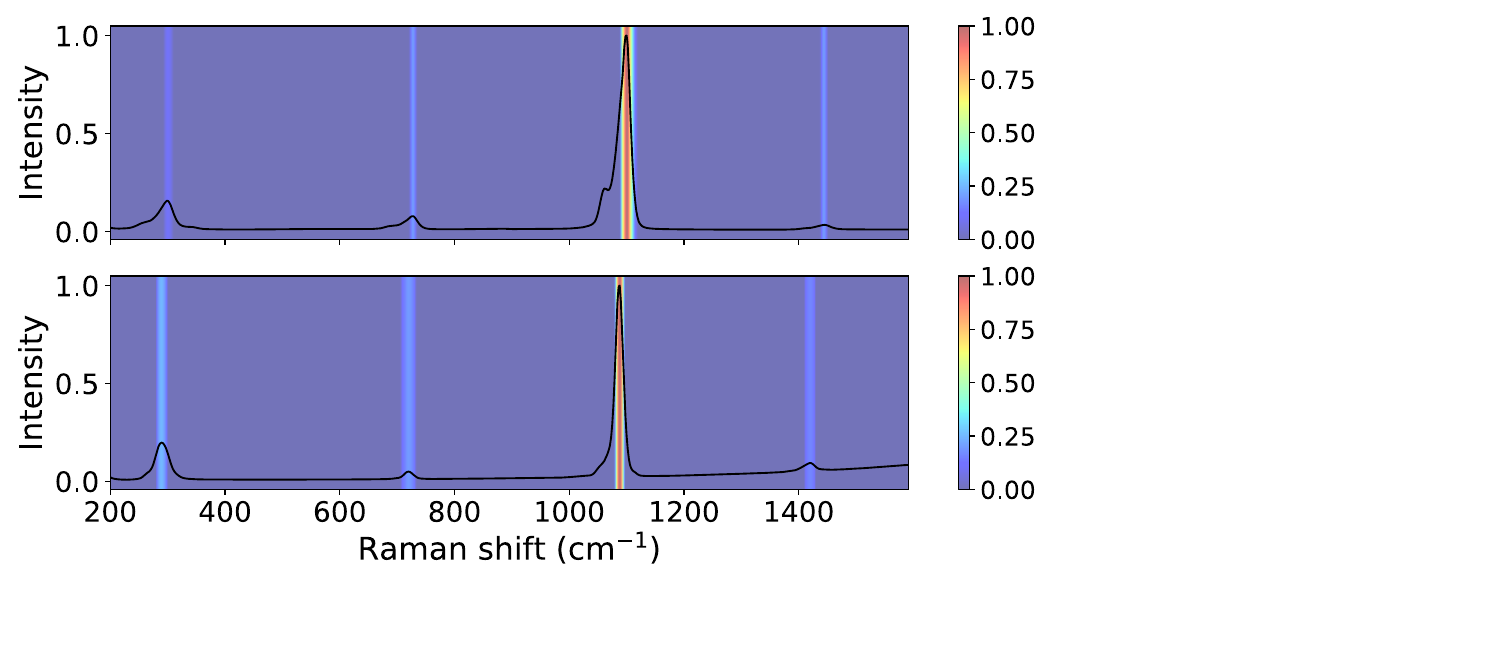}
\caption{Grad-CAM activation maps for two carbonates.
       \textit{Top}: Dolomite (\ce{CaMg(CO3)2}), correctly
       classified.  
       \textit{Bottom}: Rhodochrosite (\ce{MnCO3}),
       correctly classified.}
\label{fig:grad_cam_Dolomite_Rhodochrosite}
\end{figure}

\begin{figure}[htbp]
\centering
\includegraphics[width=0.475\textwidth,trim={0 2cm 8cm 0}]{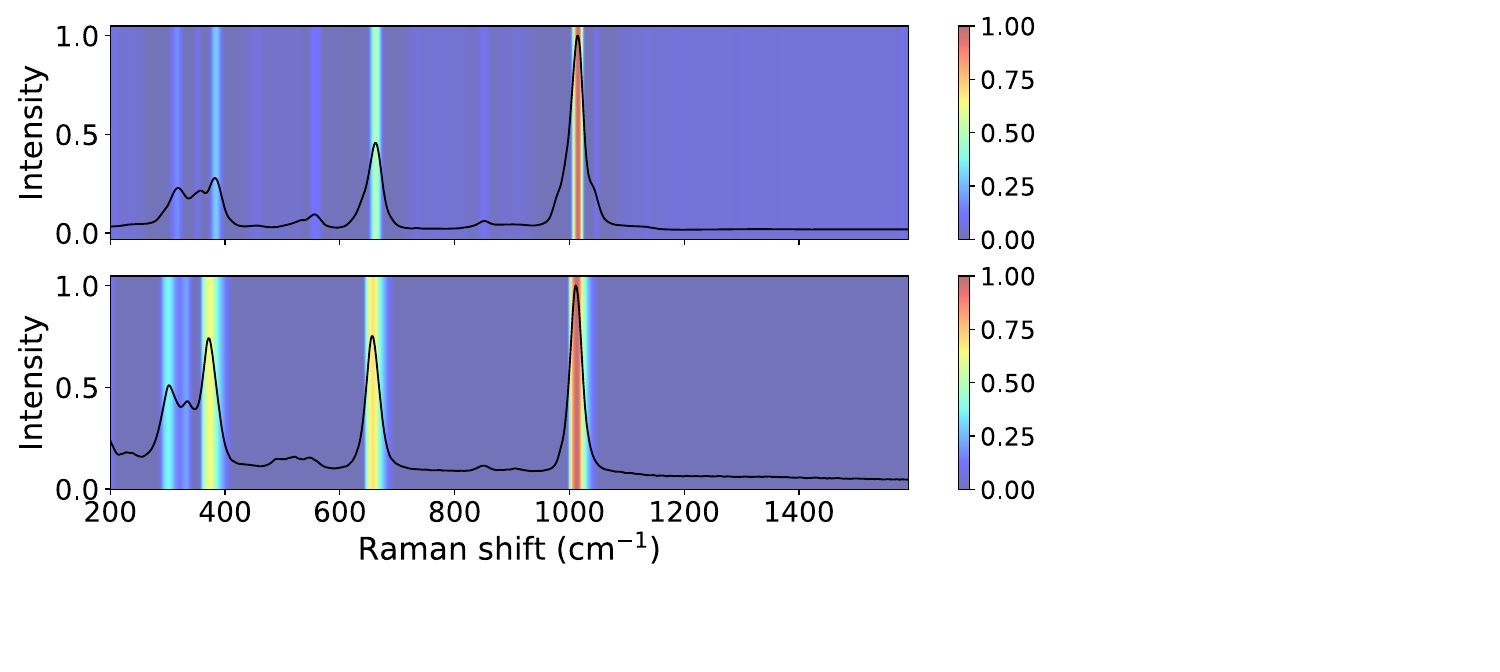}
\caption{Grad-CAM activation maps for two pyroxenes.
       \textit{Top}: Diopside (\ce{CaMgSi2O6}), misclassified as Hedenbergite.  
       \textit{Bottom}: Hedenbergite (\ce{CaFeSi2O6}), correctly classified.}
\label{fig:grad_cam_Diopside_Hedenbergite_misclassified}
\end{figure}

\begin{figure}[htbp]
\centering
\includegraphics[width=0.475\textwidth,trim={1cm 1cm 3cm 0}]{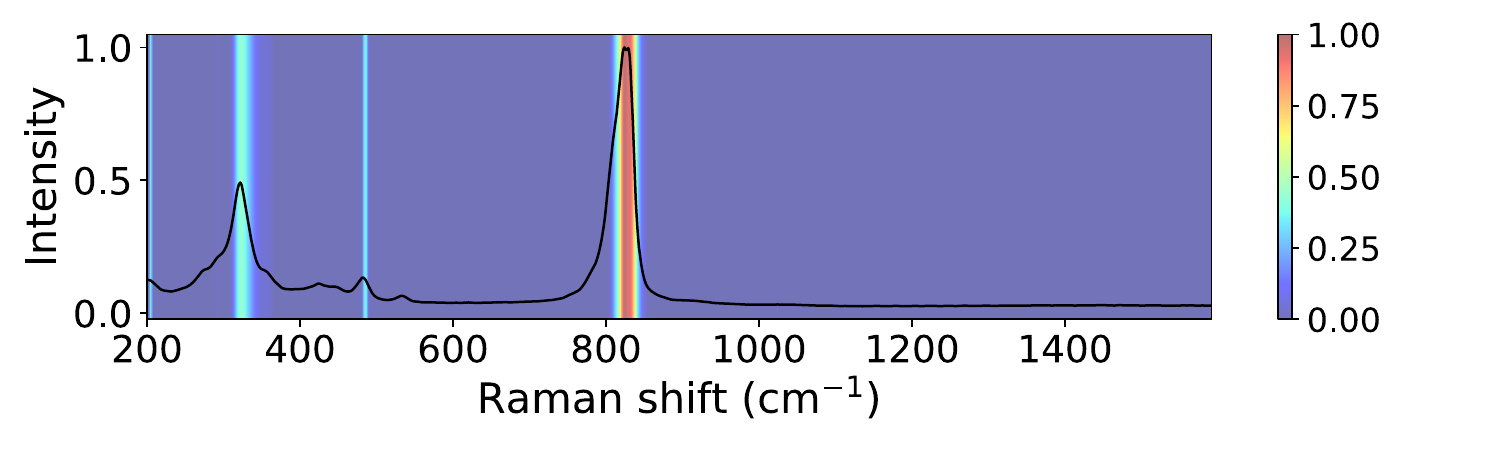}
\caption{Grad-CAM activation map. Conichalcite (\ce{CaCu(AsO4)(OH))}, correctly classified.}
\label{fig:grad_cam_Conichalcite}
\end{figure}

\begin{figure}[htbp]
\centering
\includegraphics[width=0.475\textwidth,trim={1cm 1cm 3cm 0}]{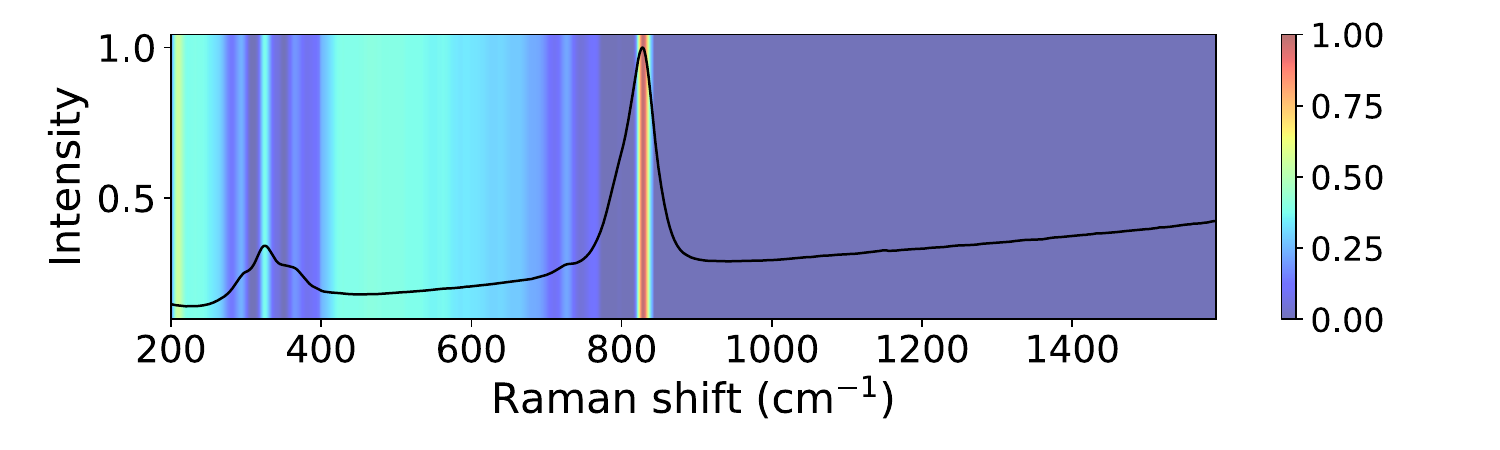}
\caption{Grad-CAM activation map. Vanadinite (\ce{Pb5(VO4)3Cl}), correctly classified.}
\label{fig:grad_cam_Vanadinite}
\end{figure}

Across the examples below the warmest regions tend to coincide with
well-known diagnostic peaks (e.g.\ the carbonate $\nu_1$ stretch at
$\sim1085\,\mathrm{cm^{-1}}$) rather than with noise or baseline structure.
This pattern indicates that CNNs focus on
spectral features that are chemically meaningful. We observe that:

\paragraph{The network correctly distinguishes similar carbonates}

Figure~\ref{fig:grad_cam_Dolomite_Rhodochrosite} contrasts two
carbonate end-members.  
For both dolomite and rhodochrosite, the hottest bars coincide with the dominant
$\nu_1$ C–O stretch \cite{Alves2023} at
$1097$ and $1085\,\mathrm{cm^{-1}}$ respectively, plus the weaker
lattice $\nu_4$ band near $300\,\mathrm{cm^{-1}}$.  
Those two regions alone suffice for the model to
distinguish Mg from Mn carbonates despite their otherwise similar
spectra.

We observe that CNNs do not merely count peaks but encodes where, to the nearest Raman
shift, those peaks reside.  
Thus even a modest $10\,\mathrm{cm^{-1}}$ displacement of the
strong $\nu_1$ line provides
enough evidence for the network to decide between dolomite or
rhodochrosite, exactly the fine-grained reasoning that a human analyst
would employ.

It is worth noting that the network used here employed a pooling size of 2, which preserves
sensitivity to such fine spectral displacements. In Section~\ref{sec:ind_bias}, we explicitly
vary the pooling size (2 vs.~64) and show that larger pooling confers robustness to
instrumental drift by introducing translational invariance, while smaller pooling retains
sensitivity to chemically meaningful shifts. The two results are therefore consistent: 
depending on the pooling hyperparameter, CNNs can be tuned either for discrimination of
closely related spectra or for robustness to noise.

\paragraph{Using peak \emph{absence} as evidence}

The Grad‑CAM maps reveal that CNNs do not merely identify single diagnostic peaks but learn combinations of peaks and their relative intensities, effectively capturing fundamental vibrational coupling patterns rather than memorizing isolated peak positions.  

For conichalcite (Figure~\ref{fig:grad_cam_Conichalcite}), the network highlights the main AsO$_4^{3-}$ $\nu_1$ stretch at $834\,\mathrm{cm^{-1}}$, alongside smaller lattice modes at $470$, $330$, and $200\,\mathrm{cm^{-1}}$ corresponding to Cu–O and O–As–O vibrations \cite{Reddy2005}.  

For vanadinite (Figure~\ref{fig:grad_cam_Vanadinite}), the network focuses on the primary VO$_4^{3-}$ $\nu_1$ stretch at $848\,\mathrm{cm^{-1}}$ and the peaks around $290$–$370\,\mathrm{cm^{-1}}$. Interestingly, it also attends to the $400$–$650\,\mathrm{cm^{-1}}$ region—the domain of conichalcite’s Cu/As bands—and uses the absence of peaks there to rule out conichalcite in favor of vanadinite.

\paragraph{A failure case: Diopside vs.~Hedenbergite}

Figure~\ref{fig:grad_cam_Diopside_Hedenbergite_misclassified} shows a
typical error when the model is asked to distinguish Mg rich diopside
(\ce{CaMgSi2O6}) from Fe rich hedenbergite (\ce{CaFeSi2O6}).  The two
Raman spectra display an almost identical three–peak motif at
$300$–$400\,\mathrm{cm^{-1}}$ and very similar band shapes
elsewhere, so the Grad-CAM heat maps illuminate the same
features in both spectra. The subtle shift of the $660\,\mathrm{cm^{-1}}$ band is not captured by the model. A larger and more compositionally diverse training set could help CNNs learn to discriminate between these closely related minerals.

\section{Understanding CNNs inductive biases for Raman spectra classification}
\label{sec:ind_bias}

While CNNs show strong empirical performance on Raman spectra, an important question remains: what mechanisms enable these models to effectively process spectroscopic data, which differs fundamentally from image data for which CNNs were originally designed~\cite{726791}? Convolution operations introduce properties such as translational equivariance, and pooling layers confer a degree of invariance — inductive biases well-suited to images but whose relevance to Raman spectra must be carefully examined.

In this section, we analyze the specific features present in Raman data and explore how the inductive bias of CNNs' architectures can be adapted to better capture the hierarchical and local patterns within spectral signals. This investigation provides deeper insights into the model’s decision-making process and guides the design of architectures tailored for spectroscopic analysis.

\subsection{Methodology: Linking Raman Characteristics with CNNs Behavior}

A key component of the underlying physical mechanism that generates the data must be considered. Raman spectroscopy, as mentioned in the Supporting Information, measures the vibrational modes of a sample. However, the vibrations are susceptible to the experimental conditions. Factors such as temperature, sample impurities or pressure can significantly distort the obtained signal (see Ferraro \textit{et al.} \cite{2003iii} for more details). In addition to alterations in the sample, the experimental equipment can also contribute to spectral distortions. The use of different lasers for the excitation of the sample and spectrometers for processing the signal can also serve as an additional noise source. These distortions in the data manifest primarily as scaling, baseline variation and peak shifts.

\begin{itemize}
	\item \textbf{Scaling:} CNNs inherently handle amplitude scaling through learned filters and, when included, normalization layers. While not inherently part of CNNs, normalization layers are commonly used to stabilize training. For SVM and KNN, the feature extraction method is also scale invariant (we only consider position of the peaks). 

	\item \textbf{Baseline variation:} CNNs, unlike traditional models, can learn to internally correct for baseline drift via convolutional filters. This eliminates the need for manual or rule-based preprocessing.

    \item \textbf{Peak shifts:} These represent the most challenging distortion. Slight shifts in Raman peak positions may result from changes in temperature, impurities, pressure, or instrument calibration. As peaks move, their associated representations change — possibly leading to misclassification. The feature extraction methods we presented in Section \ref{sec:preprocessing} for traditional ML classifiers are not invariant to shifts in peak positions. If a peak shifts enough to fall into a different interval, the features for that spectrum would change, impacting the classifier's performance. Convolutional-based models can account for translational invariance through the combination of convolutional and pooling layers. However, excessive invariance to translations can be detrimental. While smaller translations may be attributed to noise, higher translations may correspond to a different mineral. Hence, the classifier should be robust to minor translations but remain sensitive to significant ones. It is imperative to adjust the inductive bias of the CNNs to meet this requirement.
\end{itemize}

To better understand the behavior of CNNs in this context, we recall the concepts of equivariance and invariance:

\paragraph{Equivariance}

A transformation $f$ is equivariant with respect to a group of action $g$ if, for any $g$, the following relation holds:
\begin{equation}
    f(g(x)) = g(f(x)).
\end{equation}
In other words, if the input features are translated by a function $g$, then the output would also be translated by the same function. In the case of the convolution operation, defined as shown in equation \ref{eq:crossconv}, where $x$ is the input signal and $w$ is the convolution kernel, it is clear that a shift on $x$ would lead to the same shift in $S(t)$.

\begin{equation}
    S(t) = (x \ast w) (t) = \sum_{n} x(n)w(t-n).
    \label{eq:crossconv}
\end{equation}

\paragraph{Invariance}

In the case of an invariant transformation, the formulation would be:
\begin{equation}
    f(g(x)) = f(x),
\end{equation}
which means that a translation of the input features does not alter the output. 

\begin{figure*}[t]
\centering
\includegraphics[width=1.0\textwidth, trim=0 200 0 70, clip]{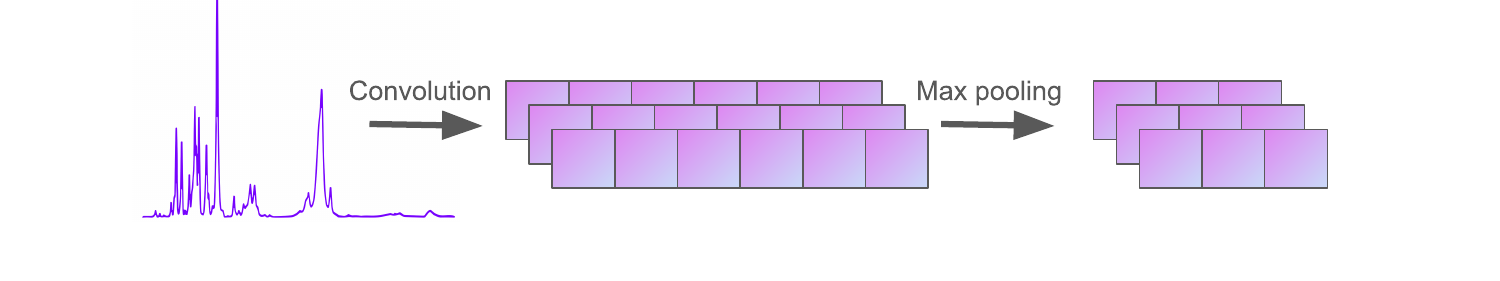}
\caption{Schematic of the max-pooling operation in a CNNs for Raman spectral analysis. The convolutional layers extract hierarchical features from the input Raman spectrum, while max pooling reduces the spatial dimensions of the feature maps, introducing local translational invariance and enhancing robustness to spectral shifts.}
\label{fig:maxpool}
\end{figure*}

For our application, a fully equivariant response—exhibiting zero invariance—is not desirable. When a peak appears at a slightly higher Raman shift, a purely equivariant system would produce a different representation, despite the underlying molecular identity remaining unchanged. In practice, such small variations often result from experimental noise or instrument calibration differences. Ideally, the model should learn representations that are invariant to these minor shifts. In convolutional neural networks, this invariance is introduced primarily through pooling layers. Specifically, max pooling is frequently used as a downsampling operation, which computes the maximum value over a window of size $m$ applied to the output of the convolutional layer. This operation, illustrated in Figure~\ref{fig:maxpool}, reduces the dimensionality of the feature maps while introducing local translational invariance. For Raman spectra, achieving an appropriate degree of translational invariance is essential: the model must be robust to small shifts caused by noise, yet still sensitive to larger shifts that may reflect different mineral species. Therefore, carefully tuning the pooling parameters is critical for adapting CNNs to Raman data classification.

\begin{figure*}[t]
\centering
\includegraphics[width=0.7\textwidth]{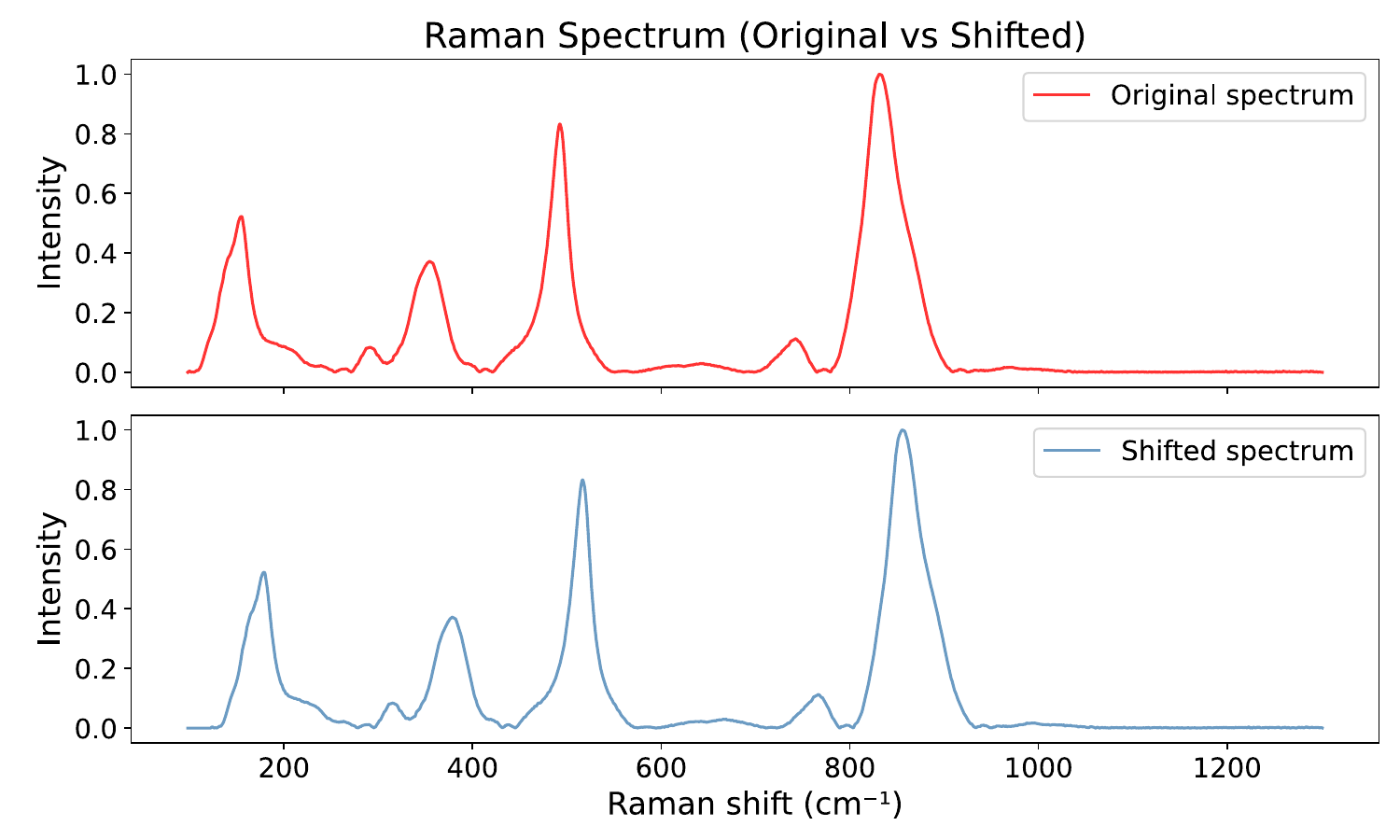}
\caption{Comparison of the original and shifted Raman spectra. The top panel shows the original spectrum, while the bottom panel displays the spectrum shifted by $30\,\mathrm{cm^{-1}}$.}
\label{fig:shifted_exp}
\end{figure*}

To explore this, we conduct experiments with artificially shifted Raman spectra, using controlled displacements of \(\pm15\) and \(\pm30\,\mathrm{cm^{-1}}\). Based on a manual inspection of the RRUFF dataset and reports that under harsh conditions, Raman spectra can exhibit shifts of up to \(30\,\mathrm{cm^{-1}}\)~\cite{liu2019temperature}, we adopt this value as a conservative upper bound on natural variation. An example of such a shifted sample is presented in Figure~\ref{fig:shifted_exp}.

\subsection{Experiments: Tuning CNNs for Partial Translational Invariance}

We examine how CNNs behave under spectral shifts by varying two key architectural parameters:

\begin{itemize}
    \item The pooling size \( m \), which controls how much each pooling layer downscales the feature map.
    \item The number of pooling layers \( n \), which affects how much invariance is accumulated across the network.
\end{itemize}

Larger \( m \) or higher \( n \) increases translational invariance — making the network more robust to shifts, but also potentially less discriminative.

We compare CNNs with low pooling size (\( m = 2 \)) and high pooling size (\( m = 64 \)), as well as networks with a shallow depth (\( n = 1 \)) and a deeper configuration (\( n = 10 \)). We report the Top-1 accuracy (fraction of spectra whose highest‑probability prediction matches the label) and the Top-3 accuracy (fraction where the true class is among the three highest‑probability predictions) for $m=2$ and $m=64$ in Table \ref{tab:pooling_param}, and the Top-1 and the Top-3 accuracy for $n=1$ and $n=5$ in Table \ref{tab:pooling_layer}.

\begin{itemize}
    \item CNNs with small pooling (low \( m \) or low \( n \)) perform well on clean data but are highly sensitive to shifts (Top-3 accuracy drops from 91\% to 16\%).
    \item CNNs with high pooling parameters show significantly better robustness (Top-3 accuracy of 57\% under \(30\,\mathrm{cm^{-1}}\) shift), albeit at a small cost in unshifted accuracy.
\end{itemize}

\begin{table*}[t]
    \captionsetup{justification=centering}
    \begin{center}
        \def\arraystretch{1.5}
        \begin{tabular}{|c|c|c|c|c|}
        \hline
        Pooling parameter & \multicolumn{2}{c|}{$m=2$} & \multicolumn{2}{c|}{$m=64$}\\
        \hline
        Accuracy (\%) &Top-1&Top-3&Top-1&Top-3\\
        \hline
        No shift&83&91&62&82\\
        \hline
        Shifted by $15\,\mathrm{cm^{-1}}$&29&51&47&65\\
        \hline
        Shifted by $30\,\mathrm{cm^{-1}}$&8&16&39&57\\
        \hline
        \end{tabular}
    \end{center}
    \caption{Top-1 and Top-3 accuracy on unshifted and shifted samples, using $m=2$ and $m=64$. Larger pooling sizes ($m=64$) show improved robustness to spectral shifts but slightly reduced accuracy on unshifted samples, reflecting a trade-off between translational invariance and resolution.}
    \label{tab:pooling_param}
\end{table*}

\begin{table*}[t]
    \captionsetup{justification=centering}
    \begin{center}
        \def\arraystretch{1.5}
        \begin{tabular}{|c|c|c|c|c|}
        \hline
        Number of pooling layers & \multicolumn{2}{c|}{$n$=1} & \multicolumn{2}{c|}{$n$=10}\\
        \hline
        Accuracy (\%) &Top-1&Top-3&Top-1&Top-3\\
        \hline
        No shift&83&91&63&80\\
        \hline
        Shifted by $15\,\mathrm{cm^{-1}}$&29&51&42&70\\
        \hline
        Shifted by $30\,\mathrm{cm^{-1}}$&6&15&46&63\\
        \hline
        \end{tabular}
    \end{center}
    \caption{Top-1 and Top-3 accuracy on unshifted and shifted samples, using $n=1$ and $n=10$ pooling layers. Increasing the number of pooling layers ($n=10$)  shows improved robustness to spectral shifts but slightly reduced accuracy on unshifted samples, emphasizing the importance of tuning inductive bias.}
    \label{tab:pooling_layer}
\end{table*}

\begin{table*}[t]
    \captionsetup{justification=centering}
    \begin{center}
        \def\arraystretch{1.5}
        \resizebox{\textwidth}{!}{%
        \begin{tabular}{|c|ccc|ccc|}
        \hline
        \# of pooling layers  & \multicolumn{3}{c|}{$n=1$}                                                                & \multicolumn{3}{c|}{$n=10$}                                                               \\ \hline
        Predicted class    & \multicolumn{1}{c|}{Top-1}         & \multicolumn{1}{c|}{Top-2}           & Top-3       & \multicolumn{1}{c|}{Top-1}         & \multicolumn{1}{c|}{Top-2}           & Top-3       \\ \hline
        No shift           & \multicolumn{1}{c|}{PyrosmaliteFe} & \multicolumn{1}{c|}{Rutile} & Diopside & \multicolumn{1}{c|}{PyrosmaliteFe} & \multicolumn{1}{c|}{Magnetite} & Grunerite \\ \hline
        Shifted by $30\,\mathrm{cm^{-1}}$ & \multicolumn{1}{c|}{Augelite}     & \multicolumn{1}{c|}{Pectolite}      & Lazulite       & \multicolumn{1}{c|}{PyrosmaliteFe} & \multicolumn{1}{c|}{Grunerite}     & Magnetite       \\ \hline 
        \end{tabular}%
        }
    \end{center}
    \caption{Top-3 predicted mineral classes for a PyrosmaliteFe sample under non-shifted and shifted conditions (\(30\,\mathrm{cm}^{-1}\)) with \(n=1\) and \(n=10\) pooling layers. Higher pooling depth preserves correct predictions under spectral shifts.}
    \label{tab:shifted_pred_layers}
\end{table*}

\begin{table*}[t]
    \captionsetup{justification=centering}
    \begin{center}
        \def\arraystretch{1.5}
        \resizebox{\textwidth}{!}{%
        \begin{tabular}{|c|ccc|ccc|}
        \hline
        Pooling parameter  & \multicolumn{3}{c|}{$m=2$}                                                                & \multicolumn{3}{c|}{$m=64$}                                                               \\ \hline
        Predicted class    & \multicolumn{1}{c|}{Top-1}         & \multicolumn{1}{c|}{Top-2}           & Top-3       & \multicolumn{1}{c|}{Top-1}         & \multicolumn{1}{c|}{Top-2}           & Top-3       \\ \hline
        No shift           & \multicolumn{1}{c|}{PyrosmaliteFe} & \multicolumn{1}{c|}{Orthoserpierite} & Paravauxite & \multicolumn{1}{c|}{PyrosmaliteFe} & \multicolumn{1}{c|}{Orthoserpierite} & Paravauxite \\ \hline
        Shifted by $30\,\mathrm{cm^{-1}}$ & \multicolumn{1}{c|}{Grunerite}     & \multicolumn{1}{c|}{Clintonite}      & Beryl       & \multicolumn{1}{c|}{PyrosmaliteFe} & \multicolumn{1}{c|}{Paravauxite}     & Beryl       \\ \hline 
        \end{tabular}%
        }
    \end{center}
    \caption{Top-3 predicted mineral classes for a PyrosmaliteFe sample under non-shifted and shifted conditions (\(30\,\mathrm{cm}^{-1}\)) with pooling sizes \(m=2\) and \(m=64\). Larger pooling improves robustness to spectral shifts, maintaining the correct class among the Top-3 predictions.}  
    \label{tab:shifted_pred}
\end{table*}

We further examine Top-3 predictions in Tables~\ref{tab:shifted_pred_layers} and~\ref{tab:shifted_pred}. CNNs with higher pooling correctly retain the true class (PyrosmaliteFe) among the Top-3 predictions for shifted inputs, while models with low pooling completely fail.

Table \ref{tab:shifted_pred_layers} represents the Top-3 predictions made by a model consisting of 1 and 10 convolutional blocks, both for the original sample and the shifted one. The model with only 1 block predicts  completely different minerals in the case of the shifted sample, while the model with 10 blocks is able to account for the shift and the Top-3 predictions are the same. Table \ref{tab:shifted_pred} represent the predictions for a model where the $m$ parameter takes values $m=2$ and $m=64$ respectively. The results obtained are consistent with the previous experiments, demonstrating that tuning $m$ can be considered as a hyperparameter that controls the degree of translational invariance. In particular, it can be tuned considering the particular instrument that is going to be used, knowing the amount of shift that can be attributed to the experimental conditions. Nonetheless, excessive pooling results in a loss of spatial information, limiting the model’s capacity to differentiate between closely related spectra.

These experiments confirm that CNNs pooling parameters (\( m \), \( n \)) act as knobs for controlling translational invariance, allowing practitioners to tailor CNNs behavior to the known variance of their spectroscopic instrumentation. This finding has direct implications for practitioners: pooling parameters can be adjusted based on the expected instrumental variance, allowing models to be tailored for specific experimental setups.

\section{Classification using semi-supervised methods}
\label{sec:semi-supervised}

CNNs typically require large amounts of labeled data, which is often expensive and time-consuming to obtain for Raman spectroscopy. In this section, we investigate two semi-supervised learning strategies to address this limitation: semi-supervised generative adversarial networks (SGAN) \cite{odena2016semisupervised} \cite{DBLP:journals/corr/SalimansGZCRC16} and contrastive learning \cite{pmlr-v119-chen20j}. Both methods aim to improve classification performance by leveraging unlabeled data.

\subsection{Methodology}
\label{sec:contrastive_learning}

\paragraph{Semi-supervised generative adversarial networks (SGAN)}

Semi-supervised generative adversarial networks (SGAN) build upon the generative adversarial network (GAN) framework, which involves two neural networks—a generator and a discriminator—trained in an adversarial setting. In the GAN framework, the generator synthesizes data that mimics the real distribution, while the discriminator learns to distinguish between real and synthetic samples. SGANs extend this framework by modifying the discriminator to output a probability distribution over \(N+1\) classes—one for each of the \(N\) real classes, plus an additional class corresponding to generated (synthetic) data. This allows the discriminator to simultaneously perform classification and real/synthetic discrimination. As a result, SGANs can leverage both labeled and unlabeled data during training, improving classification performance in low-supervision settings (see Supporting Information for details).

\begin{figure*}[t]
\centering
\includegraphics[width=0.9\textwidth]{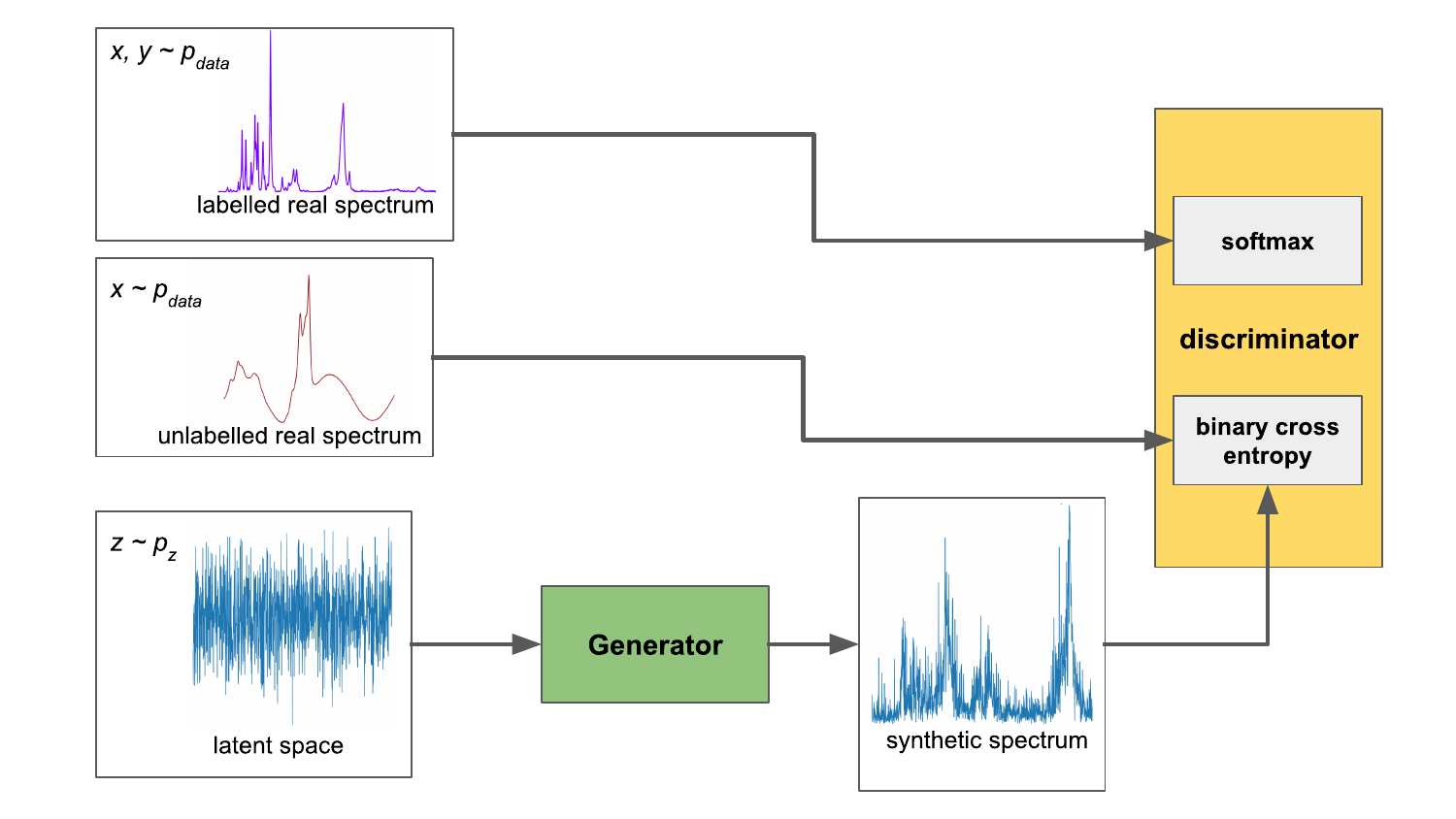}
\caption{The architecture of the SGAN applied to Raman data, featuring a 1-D CNNs-based generator that upscales latent space representations to match real spectrum dimensions, and a discriminator with an additional binary classification output to distinguish real from synthetic spectra.}
\label{fig:sgan}
\end{figure*}

In our implementation, both generator and discriminator use 1-D CNNs suited for spectral data. The generator starts from a latent vector of dimension 128 and upscales it using transposed convolutions to match the spectral shape. The discriminator uses the same base CNNs as our earlier supervised classifier (Figure \ref{fig:liuarch}), with an additional output head for binary discrimination (Figure \ref{fig:sgan}).

\paragraph{Contrastive learning}

\begin{figure*}[t]
\centering
\includegraphics[width=\textwidth]{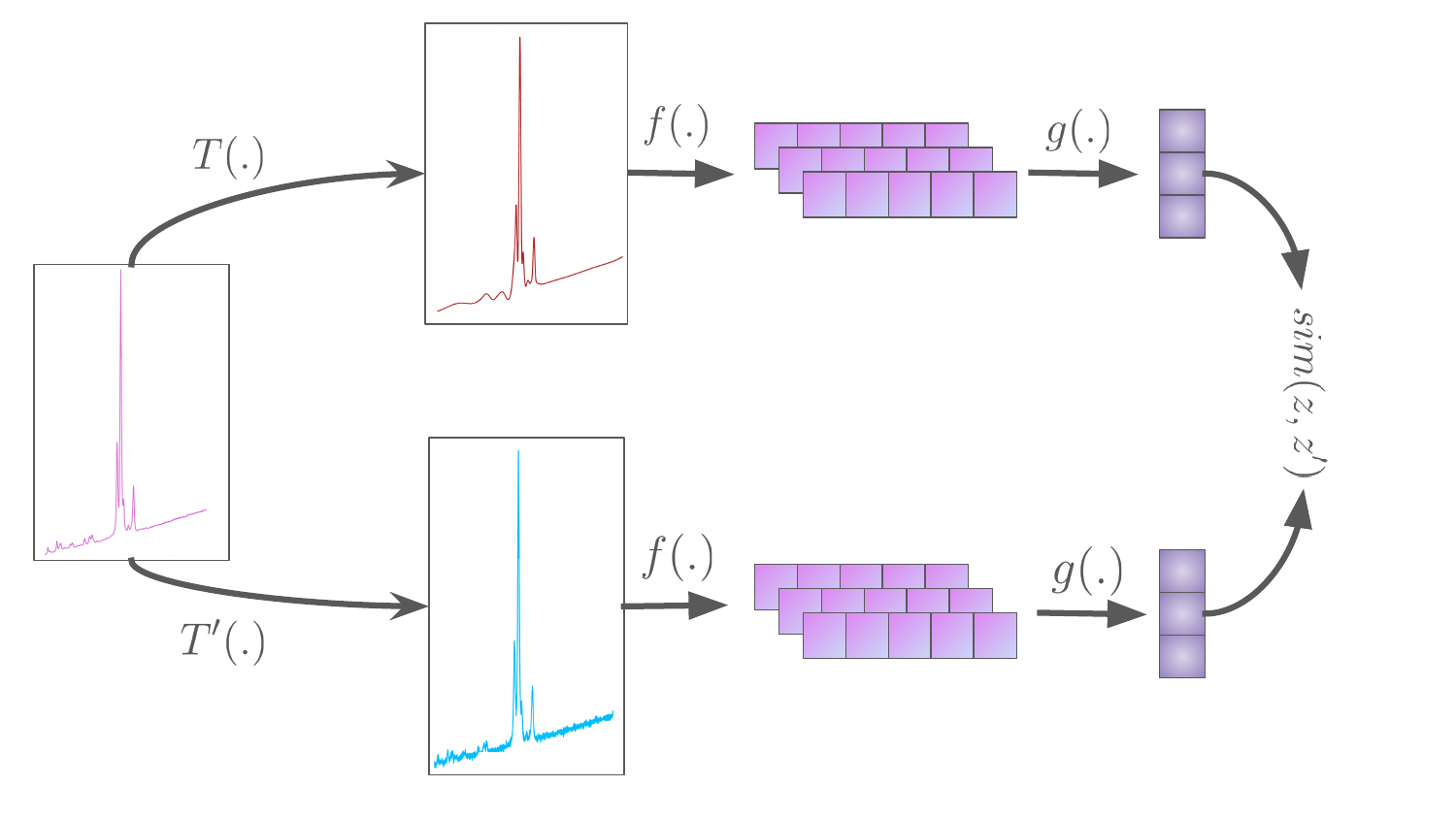}
\caption{Unsupervised contrastive learning framework applied to Raman data. The raw Raman spectrum is transformed through data augmentation functions \(T(\cdot)\) and \(T'(\cdot)\), which generate two different augmented views of the same sample. These augmented spectra are encoded into feature representations by the 1-D CNNs encoder \(f(\cdot)\). The encoded features are then passed through a projection head \(g(\cdot)\) to produce latent vectors \(z\) and \(z'\). Finally, the similarity \(\text{sim}(z, z')\) between the two latent representations is computed, encouraging the model to maximize agreement between augmented views of the same spectrum.}
\label{fig:contrasra}
\end{figure*}

Contrastive learning trains models to distinguish between similar and dissimilar data points by bringing similar pairs closer together in the representation space while pushing dissimilar pairs further apart. This is achieved by maximizing the agreement between augmented views of the same sample. We adapt the SimCLR framework \cite{pmlr-v119-chen20j} for Raman spectra using 1-D CNNs (Figure \ref{fig:contrasra}). Given a raw spectrum, we generate two stochastic augmentations using domain-specific transformations that preserve peak positions. The most general transformation we use is adding a function, such as $\tanh(\cdot)$ or a $\cos(\cdot)$ functions, with the magnitude scaled by the mean value of the spectrum. Adding a $\cos(\cdot)$ function, despite potentially introducing a new peak, can mimic the noise induced by fluorescence. Another transformation is the addition of Gaussian noise, simulating measurements from different instruments. Lastly, down-sampling and interpolating the signal, with a random selection of points in the spectrum, is used. The interpolation parameters increase the stochasticity of the method, resulting in a higher variance in the augmented spectrum. A higher weight is used for this method (that is, interpolation is used more frequently). As noted by Cheng \textit{et al.}\cite{pmlr-v119-chen20j}, unsupervised contrastive learning benefits more from data augmentation than supervised approaches, but it is crucial to ensure that the transformations cover the intra-class variance for each mineral. The encoder $f(\cdot)$ maps augmented spectra to latent features, followed by a projection head $g(\cdot)$. After unsupervised pre-training, $g(\cdot)$ is discarded, and a classifier is trained on top of frozen $f(\cdot)$ using the labeled samples (see Supporting Information for details).

\subsection{Experiments}

To simulate low-resource settings, we artificially mask the labels of a portion of the dataset, creating supervised subsets with 10\% to 50\% labeled samples. We evaluate SGAN and Contrastive learning on the raw Raman spectra and compare them to a baseline CNNs trained solely on the labeled subset.

\paragraph{SGAN: Qualitative Evaluation}

\begin{figure*}[t]
    \centering
    \begin{minipage}[t]{0.475\textwidth}
        \centering
        \includegraphics[width=0.93\textwidth]{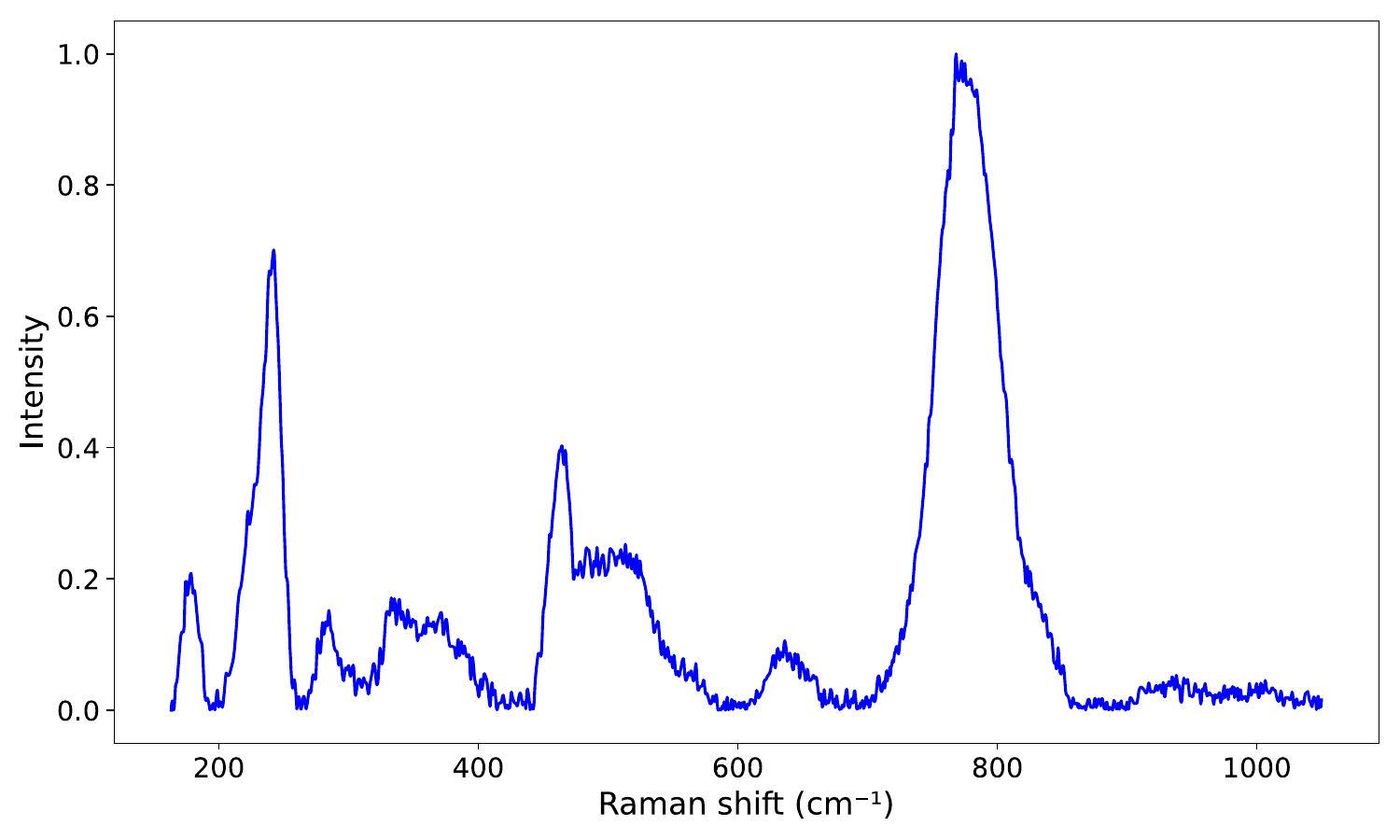}
        \caption*{(a) The real Perovskite Raman spectrum, showing characteristic peaks corresponding to the material's unique vibrational modes.}
        \label{fig:real_Pervoskite}
    \end{minipage}
    \hfill
    \begin{minipage}[t]{0.475\textwidth}
        \centering
        \includegraphics[width=\textwidth]{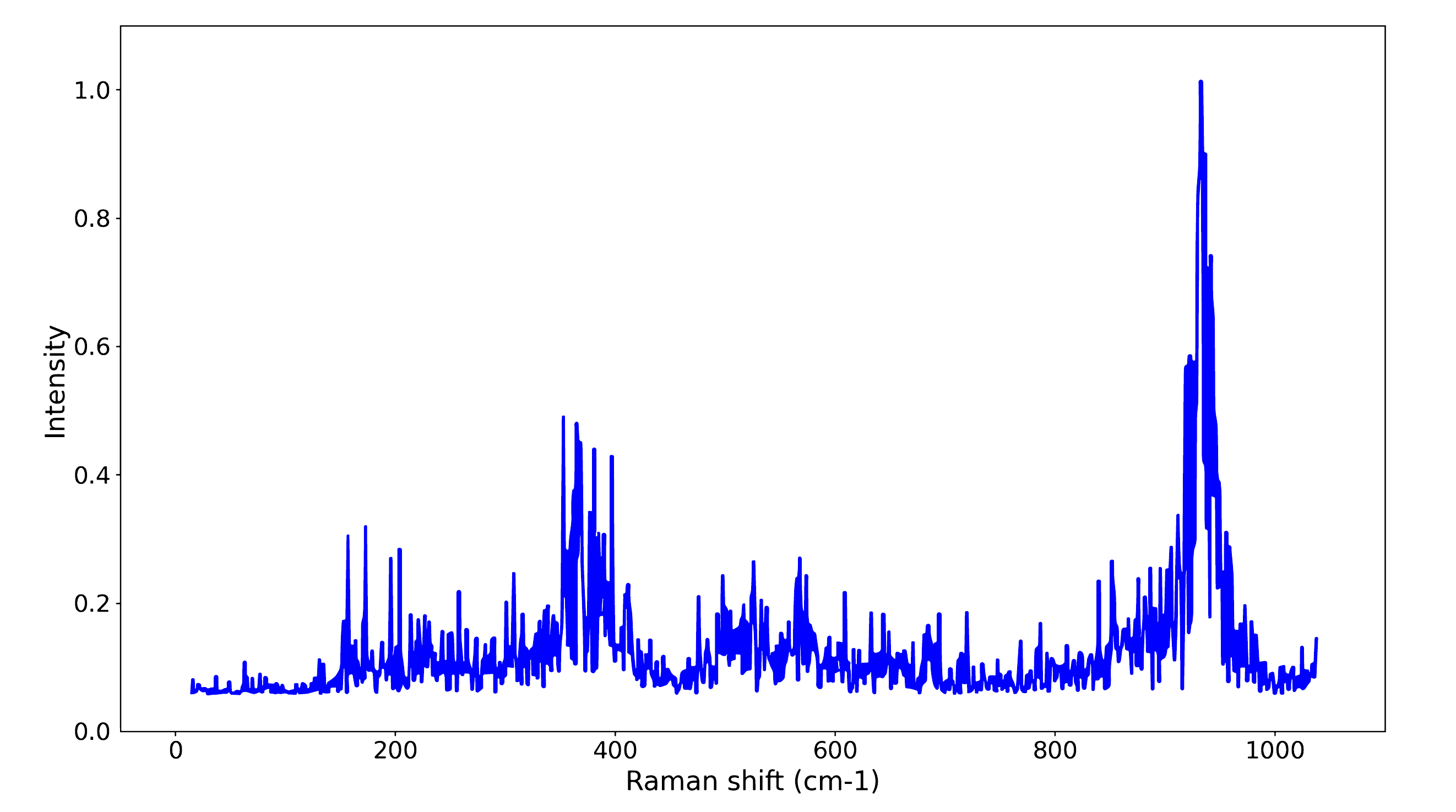}
        \caption*{(b) The synthetic Raman spectrum generated by the SGAN's generator, mimicking the general structure and noise profile of the real spectrum while preserving key spectral features.}
        \label{fig:fake_Pervoskite}
    \end{minipage}
    \caption{Comparison of real and synthetic Perovskite Raman spectra.}
    \label{fig:Pervoskite}
\end{figure*}

Figure \ref{fig:Pervoskite} compares a real Perovskite spectrum with a synthetic one generated by the SGAN. While both spectra exhibit a similar overall shape, there are noticeable differences, particularly in the noise pattern. For instance, although both spectra show a prominent peak, the peak in the generated spectrum is shifted and exhibits a distinct noise pattern compared to the real one. The primary goal is not to generate perfectly realistic samples but to enhance the performance of the discriminator/classifier through semi-supervised learning. If the generator produces samples that perfectly match the real data distribution $p(\mathbf{X})$, the decision boundaries of the discriminator remain unchanged. On the other hand, if the generated samples are \textit{complementary} to $p(\mathbf{X})$ in the feature space, this will help to obtain better decision boundaries \cite{https://doi.org/10.48550/arxiv.1705.09783}. Therefore, having generated samples that do not look exactly like the real ones is desirable in the case of SGAN, unlike traditional GAN where the principal purpose is to generate realistic samples.

The generator's task is not just to mimic real data but to create variations in the data that contribute to a richer representation of the feature space, which aids in better decision-making by the discriminator. This process can be thought of as generating additional, diverse data that helps the classifier learn more discriminative features, similar to how data augmentation methods like adding Gaussian noise can introduce variability. In both cases, the goal is to increase the robustness of the model by forcing it to adapt to these new, challenging data points.

\paragraph{SGAN: Quantitative Results}

Table \ref{tab:SGAN} shows classification accuracy on the test set using SGAN vs. a purely supervised CNNs. SGAN consistently outperforms the supervised baseline, especially when only 10\%–30\% of the data is labeled. For instance, at 10\% labeled data, SGAN improves accuracy from 42\% to 53\%. The gain diminishes as more labels become available.

However, the improvement is less than anticipated. Despite using 90\% unlabeled samples in addition to 10\% labeled ones, the expected increase in accuracy is not as substantial. This behavior can be explained by the nature of Raman spectra. Most of the literature has focused on using semi-supervised approaches like SGAN to benefit the classification problem on images. In contrast, Raman spectra exhibit different characteristics compared to image data, with low-level features having limited complexity as discussed in Section \ref{sec:low_level}. This can explain why using semi-supervised approaches in Raman spectra, although increasing the classification accuracy, has limited benefit compared to the computer-vision literature.

\begin{table*}[t]
    \captionsetup{justification=centering}
    \begin{center}
        \def\arraystretch{1.5}
        \resizebox{.8\textwidth}{!}{%
        \begin{tabular}{|l|c|c|c|c|c|}
        \hline
        Proportion of labeled samples (\%) & 10 & 20 & 30 & 40 & 50\\
        \hline
        Accuracy with only Supervised training (\%) & 42 & 58 & 65 & 69 & 76\\
        \hline
        Accuracy with Semi-Supervised training (\%) & 53 & 66 & 67 & 71 & 77\\
        \hline
        \end{tabular}
        }
    \end{center}
    \caption{Performance of fully supervised CNNs and SGAN approaches on the test set across varying proportions of labeled samples. The results show that the SGAN outperforms the fully supervised CNNs, particularly in scenarios with fewer labeled samples, highlighting the advantage of semi-supervised learning in leveraging unlabeled data.}
    \label{tab:SGAN}
\end{table*}

\paragraph{Contrastive learning: Quantitative Results}

Contrastive learning yields similar trends (Table \ref{tab:cont}). With only 10\% labeled data, accuracy improves from 42\% (supervised) to 51\% (semi-supervised). Gains diminish with more labeled data, converging to the supervised baseline. Notably, pre-training provides robust representations even when labeled examples are scarce.

In the contrastive learning framework, the pre-training phase is effective for representation learning due to the inherent characteristics of the data. Raman spectra exhibit universal features that are independent of specific classes. For instance, the fluorescence baseline is a common element in nearly all raw sample, thus a model can learn an embedding of the spectrum that accounts for this phenomenon. More broadly, similarities across all Raman spectra arise from underlying physical phenomena. This process is analogous to how Language Models learn underlying syntactic and semantic rules or patterns when pretrained with an unsupervised task \cite{https://doi.org/10.48550/arxiv.1912.09637}.

\begin{table*}[t]
\captionsetup{justification=centering}
\def\arraystretch{1.5}
\resizebox{.8\textwidth}{!}{%
\begin{tabular}{|l|c|c|c|c|c|}
\hline
Proportion of labeled samples (\%)         & 10    & 20    & 30    & 40    & 50                         \\ \hline
Accuracy with only Supervised training (\%) & 42 & 58 & 65 & 69 & 76                     \\ \hline
Accuracy with Semi-Supervised training (\%) & 51 & 65 & 69 & 75 & 76 \\ \hline
\end{tabular}
}
\caption{Performance of contrastive semi-supervised learning and fully supervised CNNs approaches on the test set across different proportions of labeled samples. The semi-supervised contrastive learning approach consistently improves classification accuracy, demonstrating its ability to effectively utilize unlabeled data to enhance performance.}

\label{tab:cont}
\end{table*}

\section{Assessing the learnability and representational efficiency of low-level features in Raman Spectra}

\label{sec:low_level}  

In this section, we investigate the nature and complexity of low-level features in Raman spectra. In image classification, convolutional neural networks (CNNs) extract hierarchical representations, where early layers capture simple features like edges, and deeper layers learn increasingly abstract patterns \cite{lecun2015deep}. While this hierarchical structure has proven effective in computer vision, it is not clear whether Raman spectra exhibit similarly complex low-level patterns.

We hypothesize that the low-level structure of Raman data enables efficient learning, such that meaningful features can be extracted from a limited number of examples. This could explain the limited gains observed with semi-supervised learning in Section~\ref{sec:semi-supervised}. To test this hypothesis, we designed two complementary experiments: first, we assess the sample efficiency of low-level feature learning using partial layer freezing in a CNNs; second, we examine the general compressibility of Raman spectra by learning sparse latent representations in an unsupervised setting. Together, these experiments allow us to evaluate both the complexity and the learnability of low-level spectral features, from both supervised and unsupervised perspectives.

\subsection{Methodology}

We structured our investigation into two targeted experiments:

\begin{enumerate}
    \item \textbf{Experiment 1: Layer Freezing} We first focus on assessing how many labeled examples are required to learn meaningful low-level features in a supervised setting. Specifically, we train the early layers of a CNNs (first two layers) using small subsets of the data, freeze them — meaning we stop updating their weights during further training — and subsequently train the remaining layers on the full dataset. Comparing this approach to an end-to-end trained CNNs helps us quantify the sample efficiency of early feature learning.

    \item \textbf{Experiment 2: Sparse Autoencoder} To complement the supervised perspective of Experiment 1, we explore whether Raman spectra can be effectively compressed into sparse latent representations in an unsupervised way. Here, we train a fully connected autoencoder to minimize reconstruction error, encouraging the model to capture the overall structure of the spectra without supervision or convolutional priors. The learned latent representation is then used as input to a downstream classifier, allowing us to evaluate how well the unsupervised features transfer to the classification task.
\end{enumerate}

\subsection{Experiments}

\paragraph{Experiment 1: Layer Freezing}

This experiment evaluates whether low-level features can be learned effectively from a limited subset of labeled data. We train the first two layers of a CNNs using subsets of increasing size (80, 200, and 848 samples), freeze them, and then train the remaining layers on the full dataset. As shown in Table~\ref{tab:freezing_layers}, performance remains stable between 200 and 848 samples, but drops from 83 \% to 76 \% with only 80 samples. These results indicate that low-level features in Raman spectra can be reliably learned from relatively few examples, highlighting the favorable sample efficiency of the data.

\begin{table*}[t]
\captionsetup{justification=centering}
\centering
\def\arraystretch{1.4}
\begin{tabular}{|c|c|c|c|}
\hline
Training Samples for Layers 1-2 & 80 & 200 & 848 \\
\hline
Test Accuracy (\%) & 76 & 82 & 83 \\
\hline
\end{tabular}
\caption{CNNs accuracy when pretraining the first two layers on subsets of increasing size. Results indicate that meaningful low-level features in Raman spectra can be learned efficiently from limited labeled data, highlighting favorable sample efficiency.}
\label{tab:freezing_layers}
\end{table*}

\paragraph{Experiment 2: Sparse Autoencoder}

While Experiment 1 demonstrates the sample efficiency of supervised learning, it does not address whether Raman spectra are inherently compressible in an unsupervised context. To investigate this, we train a fully connected autoencoder using a reconstruction loss (Equation~\ref{eq:aaloss}). The choice of reconstruction loss is critical, as it enforces the learning of a comprehensive representation of the input spectra, capturing both class-relevant and class-agnostic features. Unlike classification loss, which prioritizes discriminative characteristics, reconstruction loss ensures that the latent space encodes the full spectral information, independently of the class labels.

\begin{figure*}[t]
\centering
\includegraphics[width=0.8\textwidth]{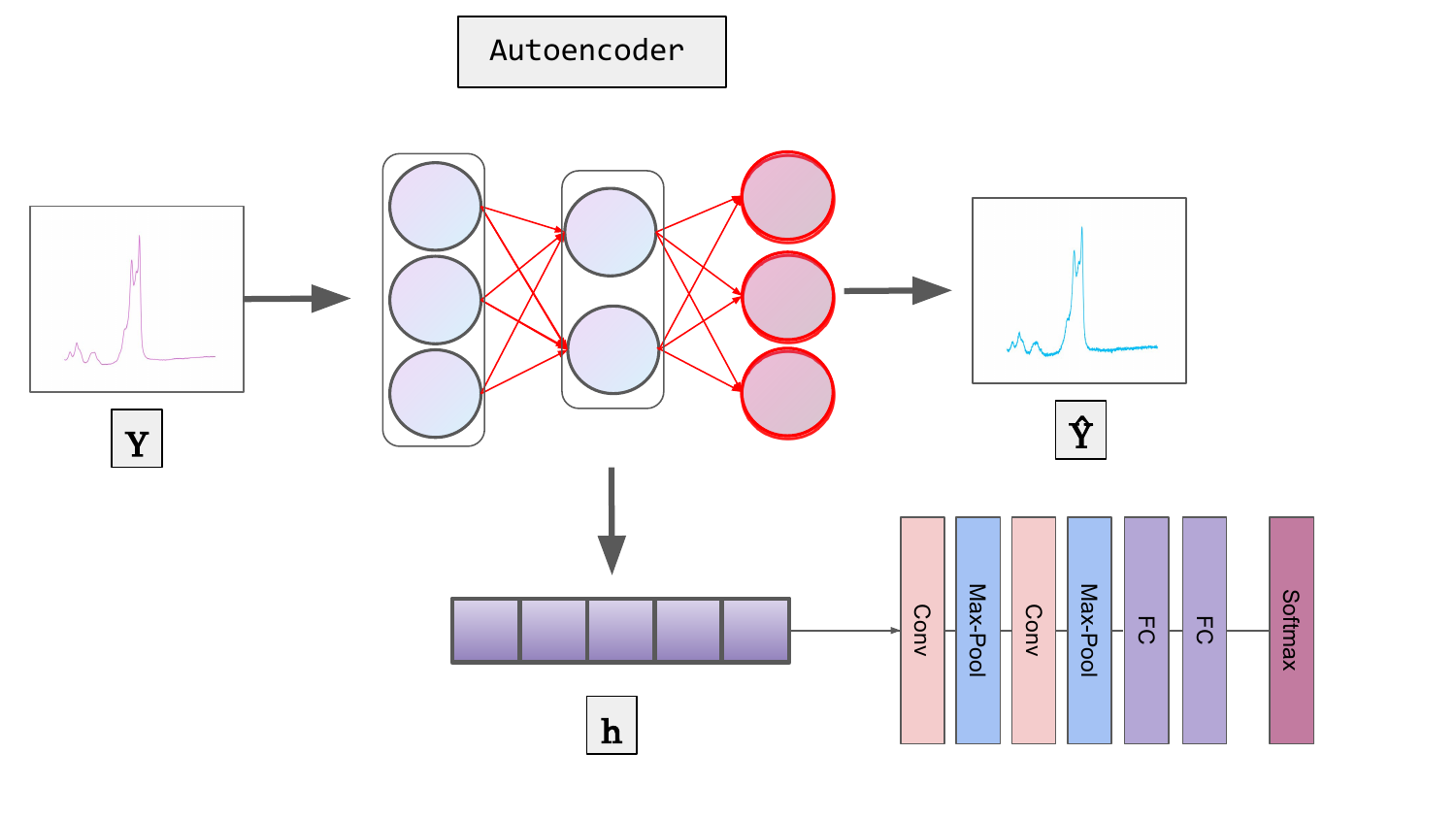}
\caption{Autoencoder experiment pipeline. The autoencoder is pretrained on a sparse reconstruction task to minimize the reconstruction error \(L(\textbf{y}, \hat{\textbf{y}})\), where \(\textbf{y}\) is the original spectrum and \(\hat{\textbf{y}}\) is the reconstructed output. This unsupervised step encourages the model to capture the full underlying structure of the spectra, independently of class labels. Low-level features are encoded in the latent representation \(\textbf{h}\) (\(\text{dim}(\textbf{h}) = 256\)). After freezing the autoencoder, \(\textbf{h}\) is used as input to train a classifier for the classification task, with the subset of data used for pretraining excluded from fine-tuning.}
\label{fig:aaexp}
\end{figure*}

In addition, by employing fully connected layers, we deliberately avoid convolutional inductive biases, allowing us to evaluate the capacity of dense architectures to model the underlying structure of Raman spectra. Following the unsupervised pretraining phase, the autoencoder layers are frozen, and the latent representations $\mathbf{h}$ are used as input features for a downstream classifier. This approach enables us to assess whether the unsupervised representations preserve sufficient discriminative information for accurate classification. To ensure methodological rigor, the dataset used for autoencoder pretraining was excluded from the subsequent classifier training. The complete experimental workflow is depicted in Figure~\ref{fig:aaexp}, illustrating the process of unsupervised autoencoder pretraining, extraction of latent spectral features, and their utilization in downstream classification.

\begin{equation}
    L(\textbf{y} , \textbf{x}) = \sum_{i=1}^{N} \left(\textbf{y}_{i}- \hat{\textbf{y}}_{i} ( \textbf{x}) \right)^{2} = \left(\textbf{y}_{i} -  \text{MLP}(\textbf{x}_{i}) \right)^{2}.
    \label{eq:aaloss}
\end{equation}

The results of the autoencoder experiment are presented in Table~\ref{tab:aar}. Despite the absence of convolutional structures and supervised training signals, the autoencoder successfully learns sparse latent representations that enable the downstream classifier to reach competitive performance. This finding complements our observations from Experiment 1: while early CNNs layers require only limited data to extract meaningful features, the autoencoder further demonstrates that Raman spectra possess an intrinsic structure that can be captured in an unsupervised and non-convolutional framework.

Notably, these results are in agreement with our previous feature engineering approach, where manually identified peak positions provided valuable classification cues (Section~\ref{sec:traditional_methods}). Here, the autoencoder extracts comparable low-level information directly from raw spectral inputs, without explicit peak detection or prior assumptions about spectral features. These findings reinforce the notion that the low-level representations in Raman spectra are both compact and discriminative, and they further support the development of efficient modeling strategies leveraging sparse representations for downstream tasks.

\begin{table*}[t]
\centering
\captionsetup{justification=centering}
\def\arraystretch{1.5}
\resizebox{0.8\textwidth}{!}{%
\begin{tabular}{|l|c|c|c|c|c|}
\hline
Proportion of labeled samples (\%)         & 10    & 20    & 30    & 40    & 50    \\ \hline
Accuracy with fully supervised CNNs (\%)     & 42    & 58    & 65    & 69    & 76    \\ \hline
Accuracy using CNNs with autoencoder features (\%)    & 43    & 53    & 58    & 66    & 67    \\ \hline
\end{tabular}
}
\caption{Comparison of fully supervised training and autoencoder-based feature extraction across varying proportions of labeled samples. The autoencoder captures informative representations of Raman spectra in an unsupervised manner, enabling competitive classification accuracy despite limited supervision.}
\label{tab:aar}
\end{table*}

\section{Classification using transfer learning}
\label{sec:transfert_learning}

\begin{table*}[hbtp]
\resizebox{0.8\textwidth}{!}{%
\def\arraystretch{1.5}
\begin{tabular}{l|c|c|c|c|}
\cline{2-5}
 & \multicolumn{1}{c|}{$c=5$} & \multicolumn{1}{c|}{$c=10$} & \multicolumn{1}{c|}{$c=15$} & \multicolumn{1}{c|}{$c=20$} \\ \hline
\multicolumn{1}{|l|}{Accuracy on \( n-c \) classes (pretraining) (\%)} & 85 & 80 & 84 & 80 \\ \hline
\multicolumn{1}{|l|}{Accuracy on \( c \) classes (fine-tuning) (\%)} & 89 & 78 & 63 & 43 \\ \hline
\end{tabular}%
}
\caption{Transfer learning results for Raman spectral classification across increasing numbers of new classes (\( c \)). High performance is maintained for small \( c \), but declines as \( c \) increases, reflecting the combined effects of reduced feature extractor diversity and increased classification complexity.}
\label{tab:transferc}
\end{table*}

Building on the insights gained from semi-supervised learning and sparse autoencoder experiments, we now explore whether representations learned by neural networks can generalize to previously unseen classes. The earlier sections demonstrated that CNNs are capable of learning robust, low-level features directly from raw Raman spectra, and that these features are sufficiently simple to be captured from a limited number of samples. These findings naturally raise the following question: can we reuse these learned features to recognize new mineral classes, without retraining the entire model from scratch?

In previous work, Liu \emph{et al.}~\cite{LIU2019175} identified a limitation of CNNs-based models for Raman classification: the need to fully retrain the network when new classes are introduced. To address this, they proposed a Siamese network architecture, which compares input spectra to reference samples. However, Siamese networks suffer from increasing inference time and storage requirements as the number of reference classes grows, since each new sample must be compared to all stored references.

As an alternative, we investigate \textbf{transfer learning}, which allows a pretrained model to be adapted to new classes with minimal retraining. This approach promises constant inference time, as new classes can be accommodated simply by retraining the final classification layer, while keeping the feature extractor frozen. Given the favorable sample efficiency of Raman spectral features and their consistency across mineral families (as shown in Section~\ref{sec:low_level}), we hypothesize that transfer learning can efficiently extend the model to new classes. Compared to Siamese networks, which require \(\mathcal{O}(k)\) comparisons to reference classes at inference time, transfer learning reduces the complexity to \(\mathcal{O}(1)\), enabling constant inference time regardless of the number of known classes.





\subsection{Methodology}

To evaluate the potential of transfer learning for Raman spectral classification, we design the following experiment:

\begin{itemize}
    \item Split the full dataset of \( n \) classes into two subsets: \( n-c \) classes for initial training and \( c \) classes reserved for fine-tuning.
    \item Train a CNNs-based model on the \( n-c \) classes, and freeze all layers of the network — that is, we stop updating their weights — except for the classification head.
    \item Replace the classification layer to accommodate the \( c \) new classes, and fine-tune only this layer for a small number of epochs.
\end{itemize}

This protocol tests whether the features learned during initial training are sufficiently general to transfer effectively to new mineral classes, without retraining the convolutional backbone. 

\subsection{Experiments}

The results of this experiment are summarized in Table~\ref{tab:transferc}. When the number of new classes \( c \) is small relative to the total number of classes \( n \), transfer learning achieves classification accuracy comparable to models trained from scratch on these new classes. However, as \( c \) increases, we simultaneously reduce the number of pretraining classes \( n - c \), which limits the diversity of the feature extractor, and increase the number of target classes, which makes the classification task inherently more complex. This combined effect leads to a degradation in performance, as shown in Table~\ref{tab:transferc}.

The training and validation curves during the fine-tuning phase for different values of \( c \) are shown in Supporting Information, Figures~S17-S20. Despite using a conservative learning rate of \( \eta = 10^{-5} \) to mitigate catastrophic forgetting, the fine-tuning process quickly converges. For small \( c \), convergence is fast and performance remains high, consistent with our findings in Section~\ref{sec:low_level} that the learned feature space is generalizable. However, as \( c \) increases, the reduced expressiveness of the frozen feature space becomes a limiting factor, and the classifier struggles to accommodate the higher number of target classes.

These results suggest that the success of transfer learning for Raman data is underpinned by the shared structural and compositional characteristics of minerals within the same family (in this context family can be substituted by another concept in the taxonomic hierarchy). Even if the model has not seen a specific mineral class during training, it can still classify it correctly if it has been exposed to related classes. This observation is coherent with our findings from the autoencoder experiment (Section~\ref{sec:low_level}), which demonstrated that low-level features generalize well across the dataset.

\section{Conclusion}

Raman spectroscopy is now deployed from planetary missions on Mars \cite{https://doi.org/10.1029/2022JE007455, Maurice2021-xe} to the investigation of hydrothermal vents on Earth \cite{Yanchilina2024, doi:10.1021/acsearthspacechem.3c00224}, and is even envisioned to facilitate the search for life on other worlds \cite{https://doi.org/10.1002/jrs.6790}. Across all of these settings, spectral interpretation faces the same core challenges: limited bandwidth, variable environmental conditions, and the need for reliable onboard autonomy. To address these challenges, we present a deployable, data-efficient deep learning framework that enables four key capabilities:

\begin{enumerate}
\item \textbf{Baseline‑free accuracy}  
1‑D CNNs surpass $k$‑nearest‑neighbors and support‑vector classifiers built on handcrafted peak features, eliminating background correction and peak picking. We also introduce physically plausible data augmentation strategies—such as controlled noise and amplitude scaling—that preserve Raman peak positions while improving generalization. All splits and preprocessing scripts are released to ensure full reproducibility.

\item \textbf{A one‑knob robustness dial}  
Tuning a single pooling parameter allows CNNs to absorb Raman shift displacements up to $30\;\mathrm{cm^{-1}}$ without degrading class resolution—providing a principled way to align model behavior with instrument stability and expected variability.

\item \textbf{Learning from unsupervised data}  
Semi‑supervised GANs and contrastive pretraining raise accuracy by up to 11\,\% with only 10\,\% of labels. Gains are modest—yet operationally crucial—due to the intrinsic information density of Raman spectra, as confirmed by inductive-bias experiments.

\item \textbf{Constant‑time transfer}  
Freezing the CNNs backbone and re‑training a softmax head adapts the model to unseen minerals at $\mathcal{O}(1)$ inference cost, outperforming Siamese and reference‑matching designs on embedded systems.
\end{enumerate}

\textbf{Reproducibility.}  
All dataset splits are publicly archived at \url{https://github.com/denizsoysal/Raman_spectra_data} to serve as a benchmark for future studies.

\textbf{Outlook.}  
As open spectral libraries expand and instrument-specific distortions become better characterized, the proposed framework—training directly on raw spectra, tuning pooling to match hardware drift, leveraging semi-supervised learning, and fine-tuning lightly for new targets—offers a scalable foundation for general-purpose Raman classifiers across minerals, organics, and biomaterials. These tools can help unlock the full potential of autonomous chemistry missions across planetary and oceanographic frontiers.

\section*{Supporting Information}
Introduction to Raman spectroscopy, detailed descriptions of the machine learning models used in this study including support vector machines, k-nearest neighbors, and convolutional neural networks, training curves and additional information on the semi-supervised and contrastive learning
methods (.PDF format).

\section*{Acknowledgements}
Deniz Soysal and Xabier García--Andrade conducted this research as part of their graduate research at KU Leuven. Renaud Detry was supported by Interne Fondsen KU Leuven/Internal Funds KU Leuven. Laura E. Rodriguez was supported by the Lunar and Planetary Institute operated by the Universities Space Research Association (LPI contribution XXXX). Pablo Sobron was supported by the SETI Institute and Impossible Sensing. Work by Laura M. Barge was carried out at the Jet Propulsion Laboratory, California Institute of Technology, under a contract with NASA (80NM0018D004). Government sponsorship acknowledged. \copyright~2025. All rights reserved.

\bibliography{references}
\end{document}